\documentclass{article}

\usepackage{arxiv}

\usepackage[utf8]{inputenc} 
\usepackage[T1]{fontenc}    
\usepackage{hyperref}       
\usepackage{url}            
\usepackage{booktabs}       
\usepackage{amsfonts}       
\usepackage{nicefrac}       
\usepackage{microtype}      
\usepackage{lipsum}		
\usepackage{graphicx}
\usepackage{doi}
\usepackage{multirow}
\usepackage{colortbl}
\usepackage{dsfont}
\usepackage{amsmath,amssymb}
\usepackage{algorithm}
\usepackage{algorithmic}

\title{A Comprehensive Review of YOLO Architectures in Computer Vision: From YOLOv1 to YOLOv8 and YOLO-NAS}

\author{ \href{https://orcid.org/0000-0001-6662-0390}{\includegraphics[scale=0.06]{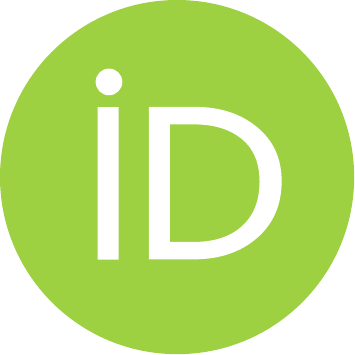}\hspace{1mm}Juan R.~Terven} \\
	Instituto Politecnico Nacional\\
        CICATA-Qro\\
	\And
	\href{https://orcid.org/0000-0002-5657-7752}{\includegraphics[scale=0.06]{orcid.pdf}\hspace{1mm}Diana M.~Cordova-Esparza} \\
	Universidad Autónoma de Querétaro\\
        Facultad de Informática\\
}

\date{}

\renewcommand{\headeright}{}
\renewcommand{\undertitle}{Published as a Journal paper at \href{https://doi.org/10.3390/make5040083}{Machine Learning and Knowledge Extraction}}
\renewcommand{\shorttitle}{Published as a Journal paper at \href{https://doi.org/10.3390/make5040083}{Machine Learning and Knowledge Extraction}}

\hypersetup{
pdftitle={A Comprehensive Review of YOLO Architectures in Computer Vision: From YOLOv1 to YOLOv8 and YOLO-NAS},
pdfsubject={cs, cs},
pdfauthor={Juan R.~Terven, Diana M.~Cordova-Esparza},
pdfkeywords={YOLO, Object Detection, Deep Learning, Computer Vision},
}

\begin{document}
\maketitle

\begin{abstract}
	YOLO has become a central real-time object detection system for robotics, driverless cars, and video monitoring applications. We present a comprehensive analysis of YOLO's evolution, examining the innovations and contributions in each iteration from the original YOLO up to YOLOv8, YOLO-NAS, and YOLO with Transformers. We start by describing the standard metrics and postprocessing; then, we discuss the major changes in network architecture and training tricks for each model. Finally, we summarize the essential lessons from YOLO's development and provide a perspective on its future, highlighting potential research directions to enhance real-time object detection systems.
\end{abstract}


\section{Introduction}
\begin{figure}[H]
  \includegraphics[width=0.98\textwidth]{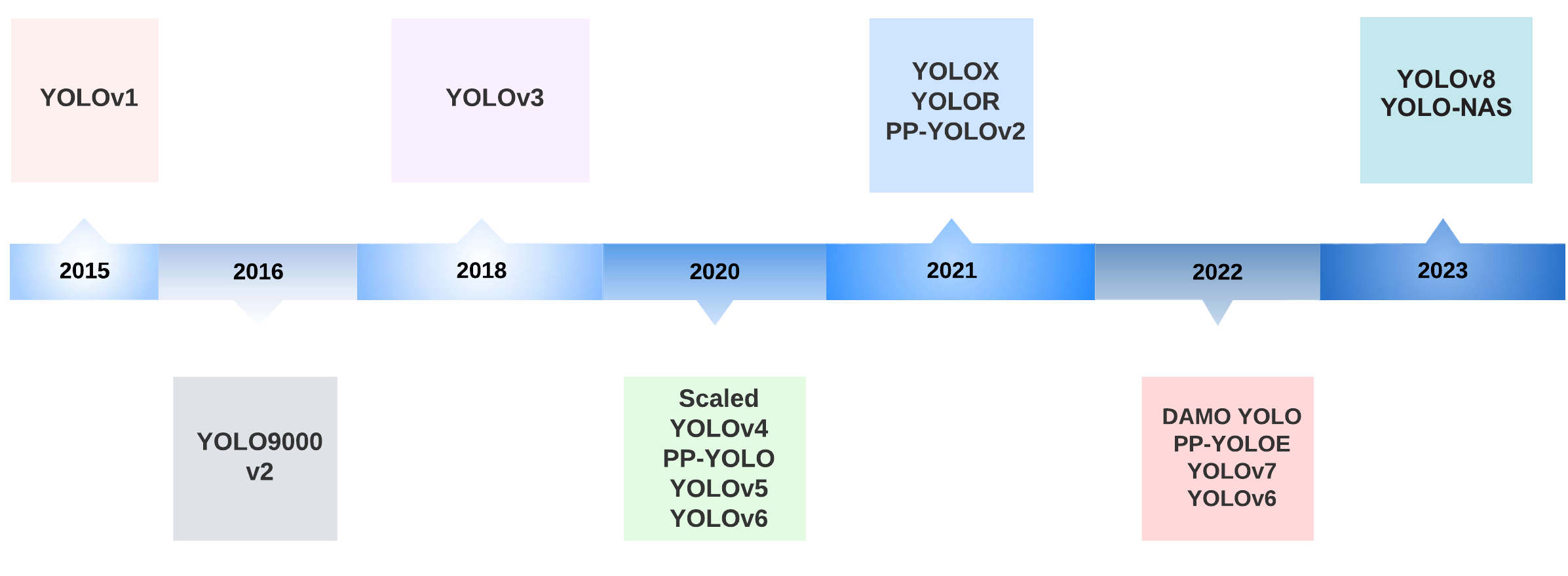}
  \caption{A timeline of YOLO versions.}
  \label{fig:yolo_timeline}
\end{figure}

Real-time object detection has emerged as a critical component in numerous applications, spanning various fields such as autonomous vehicles, robotics, video surveillance, and augmented reality. Among the different object detection algorithms, the YOLO (You Only Look Once) framework has stood out for its remarkable balance of speed and accuracy, enabling the rapid and reliable identification of objects in images. Since its inception, the YOLO family has evolved through multiple iterations, each building upon the previous versions to address limitations and enhance performance (see Figure~\ref{fig:yolo_timeline}). This paper aims to provide a comprehensive review of the YOLO framework's development, from the original YOLOv1 to the latest YOLOv8, elucidating the key innovations, differences, and improvements across each version.

In addition to the YOLO framework, the field of object detection and image processing has developed several other notable methods. Techniques such as R-CNN (Region-based Convolutional Neural Networks)~\cite{girshick2014rich} and its successors, Fast R-CNN~\cite{girshick2015fast} and Faster R-CNN~\cite{ren2015faster}, have played a pivotal role in advancing the accuracy of object detection. These methods rely on a two-stage process, where selective search generates region proposals, and convolutional neural networks classify and refine these regions. Another significant approach is the Single-Shot MultiBox Detector (SSD)\cite{liu2016ssd}, which, similar to YOLO, focuses on speed and efficiency by eliminating the need for a separate region proposal step. Additionally, methods like Mask R-NN~\cite{he2017mask} have extended capabilities to instance segmentation, enabling precise object localization and pixel-level segmentation. These developments, alongside others such as RetinaNet~\cite{lin2017focal} and EfficientDet~\cite{tan2020efficientdet}, have collectively contributed to the diverse landscape of object detection algorithms. Each method presents unique tradeoffs between speed, accuracy, and complexity, catering to different application needs and computational constraints.

Other great reviews include~\cite{bhavya2021inter,diwan2023object,hussain2023yolo}. However, the review from~\cite{bhavya2021inter} covers until YOLOv3, and~\cite{diwan2023object} covers until YOLOv4, leaving behind the most recent developments. Our paper, different from~\cite{hussain2023yolo}, shows in-depth architectures for most YOLO architectures presented and covers other variations, such as YOLOX, PP-YOLOs, YOLO with transformers, and YOLO-NAS.

This paper begins by exploring the foundational concepts and architecture of the original YOLO model, which set the stage for subsequent advances in the YOLO family. Following this, we dive into the refinements and enhancements introduced in each version, ranging from YOLOv2 to YOLOv8. These improvements encompass various aspects such as network design, loss function modifications, anchor box adaptations, and input resolution scaling. By examining these developments, we aim to offer a holistic understanding of the YOLO framework's evolution and its implications for object detection.

In addition to discussing the specific advancements of each YOLO version, the paper highlights the tradeoffs between speed and accuracy that have emerged throughout the framework's development. This underscores the importance of considering the context and requirements of specific applications when selecting the most appropriate YOLO model. Finally, we envision the future directions of the YOLO framework, touching upon potential avenues for further research and development that will shape the ongoing progress of real-time object detection systems.

\section{YOLO Applications Across Diverse Fields}

YOLO's real-time object detection capabilities have been invaluable in autonomous vehicle systems, enabling quick identification and tracking of various objects such as vehicles, pedestrians \cite{lan2018pedestrian,hsu2021adaptive}, bicycles, and other obstacles \cite{autonomous_vehicles_benjumea2021yolo,autonomous_vehicles_dazlee2022object,autonomous_vehicles_liang2022edge,li2021detection}. These capabilities have been applied in numerous fields, including action recognition \cite{action_recog_shinde2018yolo} in video sequences for surveillance \cite{surveillance_ashraf2022weapons}, sports analysis \cite{zheng2022video}, and human-computer interaction \cite{ma2021fer}.

YOLO models have been used in agriculture to detect and classify crops \cite{agriculture_tian2019apple,agriculture_wu2020using}, pests, and diseases \cite{agriculture_lippi2021yolo}, assisting in precision agriculture techniques and automating farming processes. They have also been adapted for face detection tasks in biometrics, security, and facial recognition systems \cite{yang2018real,chen2021yolo}.

In the medical field, YOLO has been employed for cancer detection \cite{al2018simultaneous,nie2019automatic}, skin segmentation \cite{unver2019skin}, and pill identification \cite{tan2021comparison}, leading to improved diagnostic accuracy and more efficient treatment processes. In remote sensing, it has been used for object detection and classification in satellite and aerial imagery, aiding in land use mapping, urban planning, and environmental monitoring \cite{cheng2021small,pham2020yolo,qing2021improved,zakria2022multiscale}.

Security systems have integrated YOLO models for real-time monitoring and analysis of video feeds, allowing rapid detection of suspicious activities \cite{kumar2021real}, social distancing, and face mask detection \cite{bhambani2020real}. The models have also been applied in surface inspection to detect defects and anomalies, enhancing quality control in manufacturing and production processes \cite{li2018real,ukhwah2019asphalt,du2021pavement}.

\begin{figure}[H]
  \includegraphics[width=\textwidth]{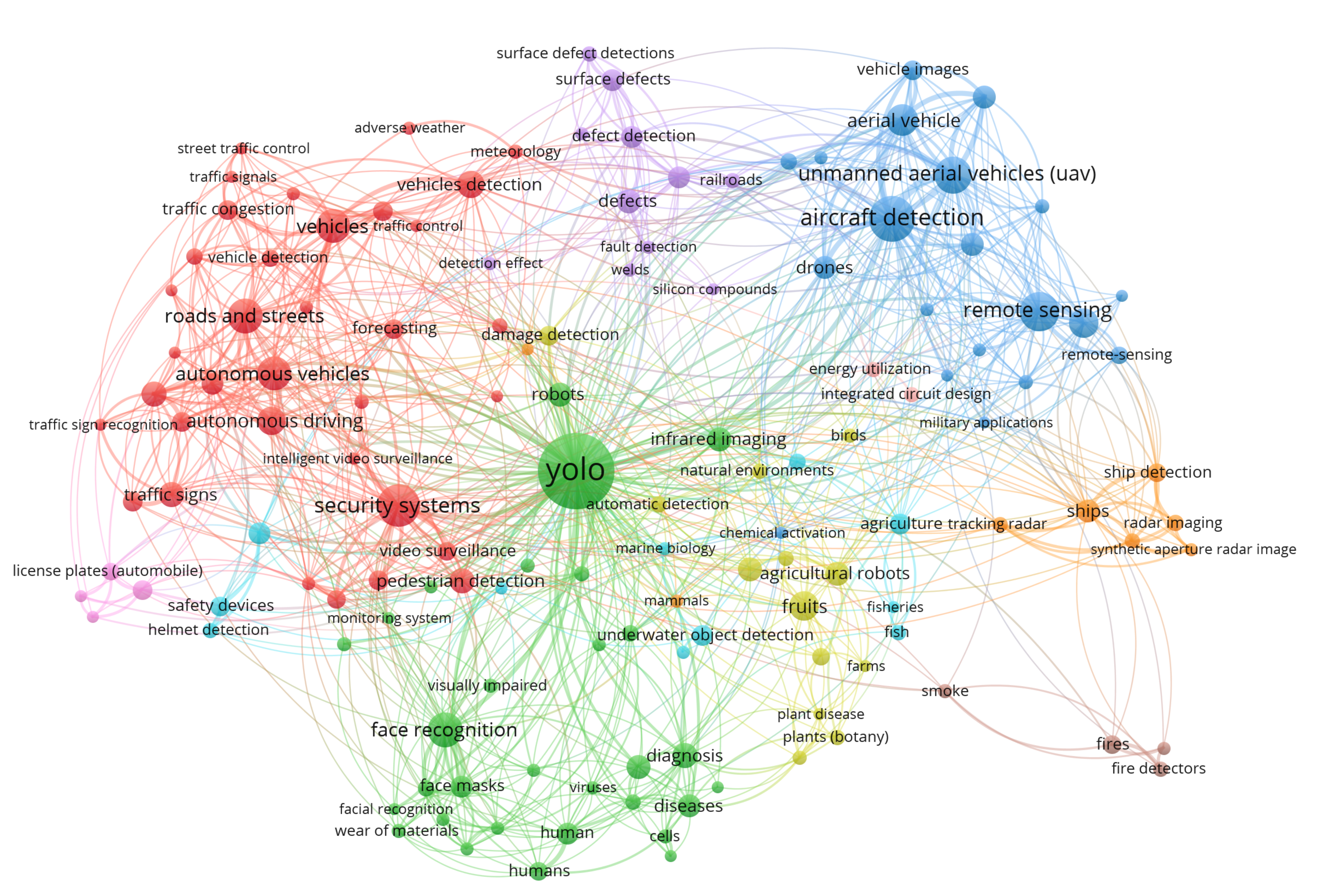}
  \caption{Bibliometric network visualization of the main YOLO Applications created with~\cite{VOSviewer_Visualizing_Scientific_Landscapes_2023}.}
  \label{fig:yolo_applications}
\end{figure}

In traffic applications, YOLO models have been utilized for tasks such as license plate detection~\cite{chen2019automatic} and traffic sign recognition~\cite{dewi2022deep}, contributing to developing intelligent transportation systems and traffic management solutions. They have been employed in wildlife detection and monitoring to identify endangered species for biodiversity conservation and ecosystem management~\cite{roy2023wildect}. Lastly, YOLO has been widely used in robotic applications~\cite{kulik2020experiments,dos2019mobile} and object detection from drones~\cite{sahin2021yolodrone,chen2023yolo}. Figure~\ref{fig:yolo_applications} shows a bibliometric network visualization of all the papers found in Scopus with the word YOLO in the title and filtered by object detection keyword. Then, we manually filtered all the papers related to applications.

\section{Object Detection Metrics and Non-Maximum Suppression (NMS)}
The Average Precision (AP), traditionally called \textit{Mean Average Precision} (mAP), is the commonly used metric for evaluating the performance of object detection models. It measures the average precision across all categories, providing a single value to compare different models. The COCO dataset makes no distinction between AP and mAP. In the rest of this paper, we will refer to this metric as AP.

In YOLOv1 and YOLOv2, the dataset utilized for training and benchmarking was PASCAL VOC 2007, and VOC 2012 \cite{everingham2010pascal}. However, from YOLOv3 onwards, the dataset used is Microsoft COCO (Common Objects in Context) \cite{lin2014microsoft}. The AP is calculated differently for these datasets. The following sections will discuss the rationale behind AP and explain how it is computed.  

\subsection{How AP works?}
The AP metric is based on precision-recall metrics, handling multiple object categories, and defining a positive prediction using Intersection over Union (IoU).

\textbf{Precision and Recall}: Precision measures the accuracy of the model's positive predictions, while recall measures the proportion of actual positive cases that the model correctly identifies. There is often a trade-off between precision and recall; for example, increasing the number of detected objects (higher recall) can result in more false positives (lower precision). To account for this trade-off, the AP metric incorporates the precision-recall curve that plots precision against recall for different confidence thresholds. This metric provides a balanced assessment of precision and recall by considering the area under the precision-recall curve.

\textbf{Handling multiple object categories}: Object detection models must identify and localize multiple object categories in an image. The AP metric addresses this by calculating each category's average precision (AP) separately and then taking the mean of these APs across all categories (that is why it is also called mean average precision). This approach ensures that the model's performance is evaluated for each category individually, providing a more comprehensive assessment of the model's overall performance.

\begin{figure}[ht]
  \centering
  \includegraphics[width=13cm]{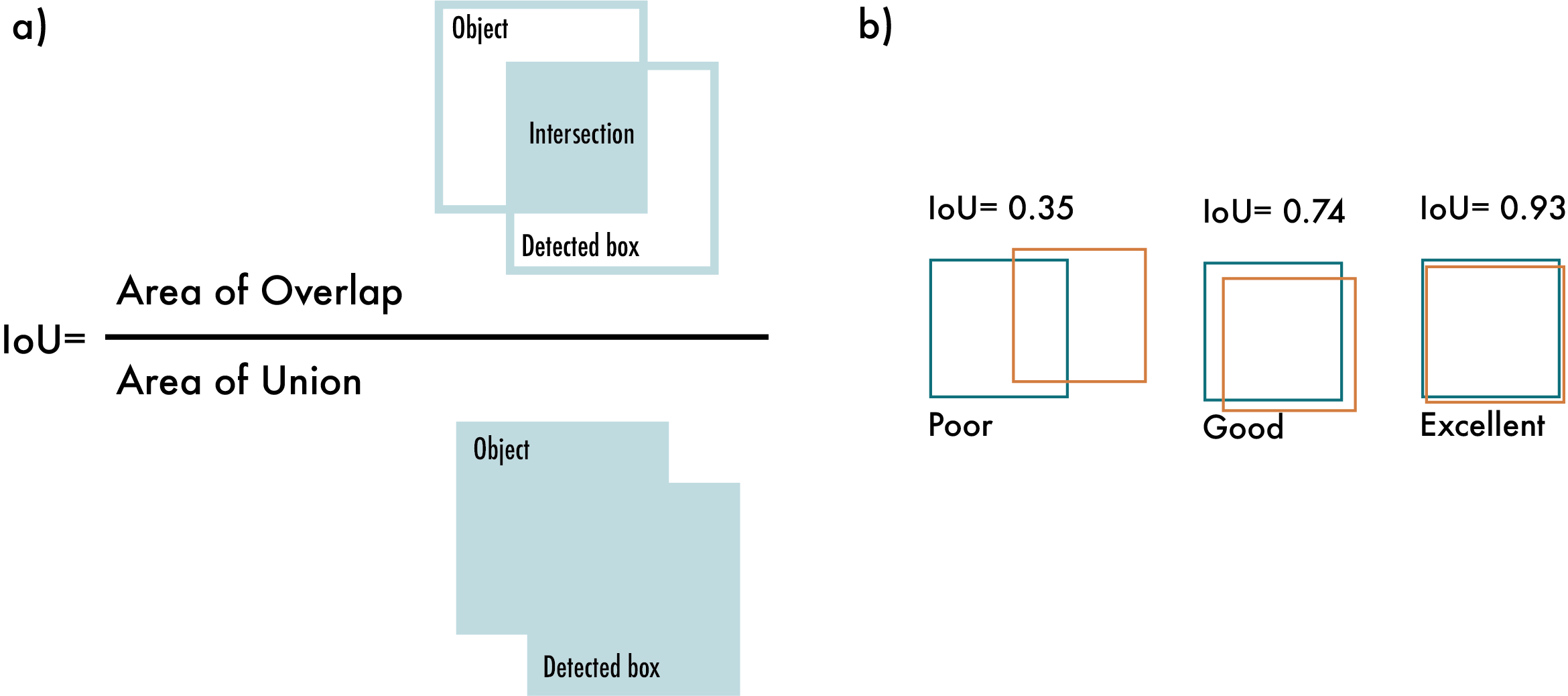}
  \caption{Intersection over Union (IoU).  a) The IoU is calculated by dividing the intersection of the two boxes by the union of the boxes; b) examples of three different IoU values for different box locations.}
  \label{fig:IoU}
\end{figure}

\textbf{Intersection over Union}: Object detection aims to accurately localize objects in images by predicting bounding boxes. The AP metric incorporates the Intersection over Union (IoU) measure to assess the quality of the predicted bounding boxes. IoU is the ratio of the intersection area to the union area of the predicted bounding box and the ground truth bounding box (see Figure \ref{fig:IoU}). It measures the overlap between the ground truth and predicted bounding boxes. The COCO benchmark considers multiple IoU thresholds to evaluate the model's performance at different levels of localization accuracy.

\subsection{Computing AP}
The AP is computed differently in the VOC and in the COCO datasets. In this section, we describe how it is computed on each dataset.

\subsubsection*{VOC Dataset}  
 This dataset includes 20 object categories. To compute the AP in VOC, we follow the next steps:
 \begin{enumerate}
     \item For each category, calculate the precision-recall curve by varying the confidence threshold of the model's predictions.
     \item Calculate each category's average precision (AP) using an interpolated 11-point sampling of the precision-recall curve.
     \item Compute the final average precision (AP) by taking the mean of the APs across all 20 categories.
 \end{enumerate}

\subsubsection*{Microsoft COCO Dataset}
This dataset includes 80 object categories and uses a more complex method for calculating AP. Instead of using an 11-point interpolation, it uses a 101-point interpolation, i.e., it computes the precision for 101 recall thresholds from 0 to 1 in increments of 0.01. Also, the AP is obtained by averaging over multiple IoU values instead of just one, except for a common AP metric called $AP_{50}$, which is the AP for a single IoU threshold of 0.5. 
The steps for computing AP in COCO are the following:
\begin{enumerate}
    \item For each category, calculate the precision-recall curve by varying the confidence threshold of the model's predictions.
    \item Compute each category's average precision (AP) using 101-recall thresholds.
    \item Calculate AP at different Intersection over Union (IoU) thresholds, typically from 0.5 to 0.95 with a step size of 0.05. A higher IoU threshold requires a more accurate prediction to be considered a true positive.
    \item For each IoU threshold, take the mean of the APs across all 80 categories.
    \item Finally, compute the overall AP by averaging the AP values calculated at each IoU threshold.
\end{enumerate}

The differences in AP calculation make it hard to directly compare the performance of object detection models across the two datasets. The current standard uses the COCO AP due to its more fine-grained evaluation of how well a model performs at different IoU thresholds.

\subsection{Non-Maximum Suppression (NMS)}
\begin{figure}[ht]
  \centering
  \includegraphics[width=12cm]{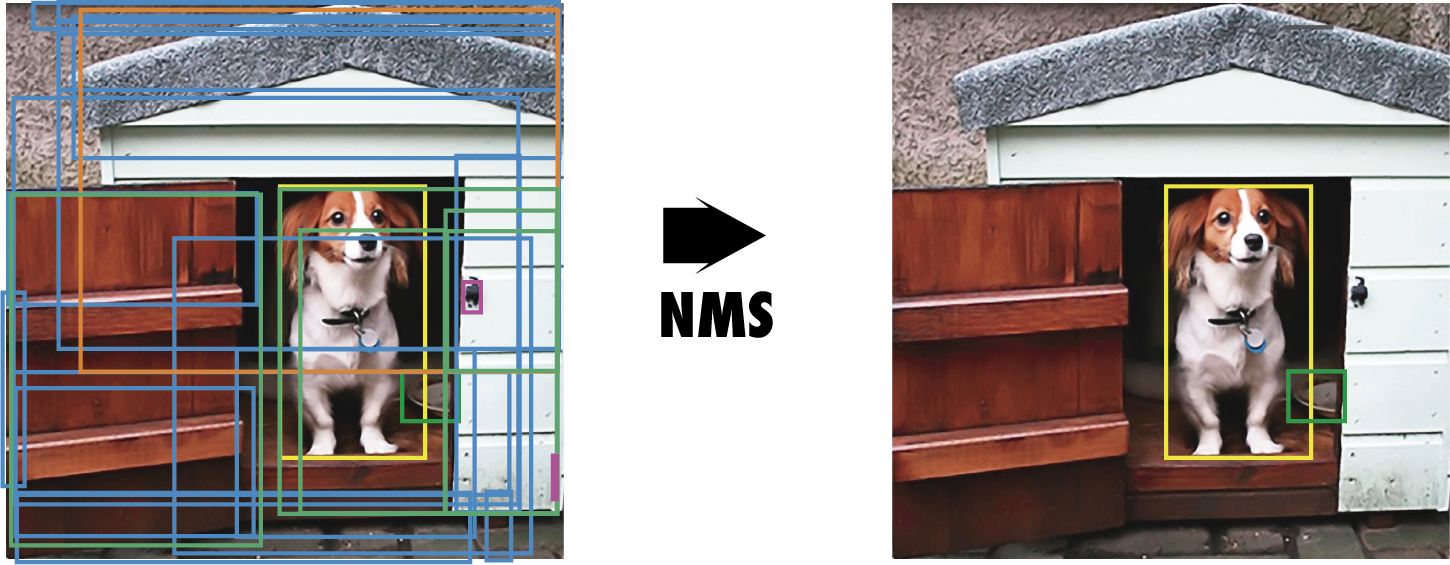}
  \caption{Non-Maximum Suppression (NMS). a) Shows the typical output of an object detection model containing multiple overlapping boxes. b) Shows the output after NMS.}
  \label{fig:nms}
\end{figure}

Non-Maximum Suppression (NMS) is a post-processing technique used in object detection algorithms to reduce the number of overlapping bounding boxes and improve the overall detection quality. Object detection algorithms typically generate multiple bounding boxes around the same object with different confidence scores. NMS filters out redundant and irrelevant bounding boxes, keeping only the most accurate ones. Algorithm \ref{alg:nms} describes the procedure. Figure \ref{fig:nms} shows the typical output of an object detection model containing multiple overlapping bounding boxes and the output after NMS.

\begin{algorithm}
\caption{Non-Maximum Suppression Algorithm}
\label{alg:nms}
\begin{algorithmic}[1]
    \REQUIRE Set of predicted bounding boxes $B$, confidence scores $S$, IoU threshold $\tau$, confidence threshold $T$
    \ENSURE Set of filtered bounding boxes $F$
    \STATE $F \gets \emptyset$
    \STATE Filter the boxes: $B \gets \{b \in B \mid S(b) \geq T\}$
    \STATE Sort the boxes $B$ by their confidence scores in descending order
    \WHILE{$B \neq \emptyset$}
        \STATE Select the box $b$ with the highest confidence score
        \STATE Add $b$ to the set of final boxes $F$: $F \gets F \cup \{b\}$
        \STATE Remove $b$ from the set of boxes $B$: $B \gets B - \{b\}$
        \FORALL{remaining boxes $r$ in $B$}
            \STATE Calculate the IoU between $b$ and $r$: $iou \gets IoU(b, r)$
            \IF{$iou \geq \tau$}
                 \STATE Remove $r$ from the set of boxes $B$: $B \gets B - \{r\}$
            \ENDIF
        \ENDFOR
    \ENDWHILE
\end{algorithmic}
\end{algorithm}

We are ready to start describing the different YOLO models.

\section{YOLO: You Only Look Once}
YOLO by Joseph Redmon et al. was published in CVPR 2016 \cite{redmon2016you}. It presented for the first time a real-time end-to-end approach for object detection. The name YOLO stands for "You Only Look Once," referring to the fact that it was able to accomplish the detection task with a single pass of the network, as opposed to previous approaches that either used sliding windows followed by a classifier that needed to run hundreds or thousands of times per image or the more advanced methods that divided the task into two-steps, where the first step detects possible regions with objects or \emph{regions proposals} and the second step run a classifier on the proposals. Also, YOLO used a more straightforward output based only on regression to predict the detection outputs as opposed to Fast R-CNN \cite{girshick2015fast} that used two separate outputs, a classification for the probabilities and a regression for the boxes coordinates. 

\subsection{How YOLOv1 works?}
YOLOv1 unified the object detection steps by detecting all the bounding boxes simultaneously. To accomplish this, YOLO divides the input image into a $S \times S$ grid and predicts $B$ bounding boxes of the same class, along with its confidence for $C$ different classes per grid element. Each bounding box prediction consists of five values: $Pc, bx, by, bh, bw$ where $Pc$ is the confidence score for the box that reflects how confident the model is that the box contains an object and how accurate the box is. The $bx$ and $by$ coordinates are the centers of the box relative to the grid cell, and $bh$ and $bw$ are the height and width of the box relative to the full image.  
The output of YOLO is a tensor of $S \times S \times (B \times 5 + C)$ optionally followed by non-maximum suppression (NMS) to remove duplicate detections. 

In the original YOLO paper, the authors used the PASCAL VOC dataset \cite{everingham2010pascal} that contains 20 classes ($C=20$); a grid of $7\times7$ ($S=7$) and at most $2$ classes per grid element ($B=2$), giving a $7 \times 7 \times 30$ output prediction.

\begin{figure}[ht]
  \centering
  \includegraphics[width=\linewidth]{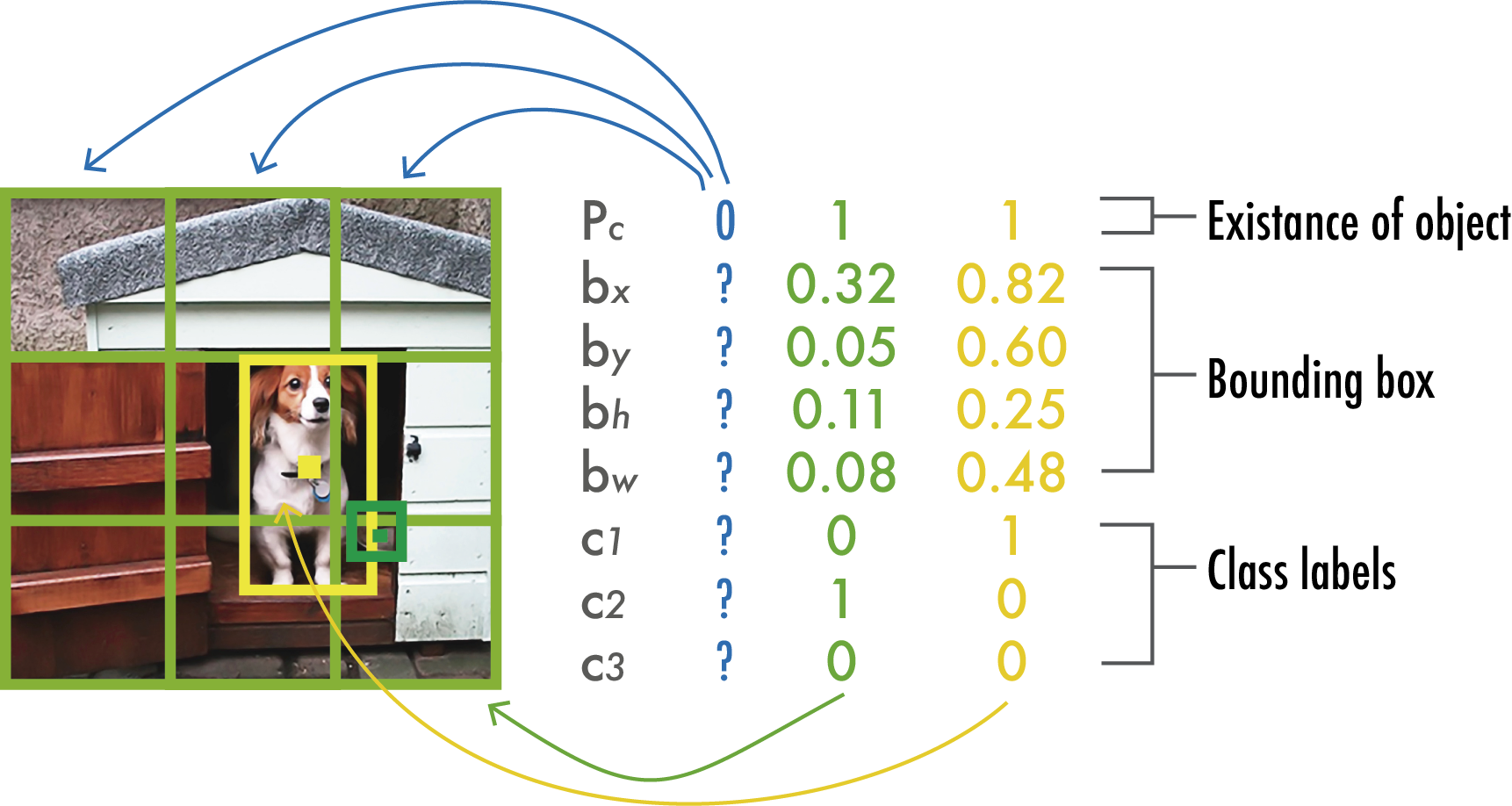}
  \caption{YOLO output prediction. The figure depicts a simplified YOLO model with a three-by-three grid, three classes, and a single class prediction per grid element to produce a vector of eight values.}
  \label{fig:yolo_out_pred}
\end{figure}

Figure \ref{fig:yolo_out_pred} shows a simplified output vector considering a three-by-three grid, three classes, and a single class per grid for eight values. In this simplified case, the output of YOLO would be $3\times3\times 8$.

YOLOv1 achieved an average precision (AP) of 63.4 on the PASCAL VOC2007 dataset.

\subsection{YOLOv1 Architecture}
YOLOv1 architecture comprises 24 convolutional layers followed by two fully-connected layers that predict the bounding box coordinates and probabilities. All layers used leaky rectified linear unit activations \cite{maas2013rectifier} except for the last one that used a linear activation function.  
Inspired by GoogLeNet \cite{szegedy2015going} and Network in Network \cite{lin2013network}, YOLO uses $1\times1$ convolutional layers to reduce the number of feature maps and keep the number of parameters relatively low. As activation layers, Table \ref{tab:yolov1} describes the YOLOv1 architecture. 
The authors also introduced a lighter model called Fast YOLO, composed of nine convolutional layers.

\begin{table}
  \centering
  \caption{YOLO Architecture. The architecture comprises 24 convolutional layers combining $3\times3$ convolutions with $1\times1$ convolutions for channel reduction. The output is a fully connected layer that generates a grid of $7\times7$ with 30 values for each grid cell to accommodate ten bounding box coordinates (2 boxes) with 20 categories.}
  \label{tab:yolov1}
  \begin{tabular}{lllll}
   \hline
     & Type & Filters & Size/Stride & Output\\ 
    \hline
        & Conv     & 64  & $7\times7$ / 2 & $224\times224$ \\
        & Max Pool &     & $2\times2$ / 2 & $112\times112$ \\
        & Conv     & 192 & $3\times3$ / 1 & $112\times112$ \\
        & Max Pool &     & $2\times2$ / 2 & $56\times56$ \\
     \toprule
     \multirow{2}{1cm}{$1\times$} &  Conv     & 128 & $1\times1$ / 1 & $56\times56$ \\   
        &Conv     & 256 & $3\times3$ / 1 & $56\times56$ \\  
        & Conv     & 256 & $1\times1$ / 1 & $56\times56$ \\
        & Conv     & 512 & $3\times3$ / 1 & $56\times56$ \\
        & Max Pool &     & $2\times2$ / 2 & $28\times28$ \\
       \toprule
         \multirow{2}{1cm}{$4\times$} & Conv     & 256 & $1\times1$ / 1 & $28\times28$ \\
        &Conv     & 512 & $3\times3$ / 1 & $28\times28$ \\
        \bottomrule
        & Conv     & 512 & $1\times1$ / 1 & $28\times28$ \\
        & Conv     & 1024 & $3\times3$ / 1 & $28\times28$ \\
        & Max Pool &     & $2\times2$ / 2 & $14\times14$ \\
    \toprule
    \multirow{2}{1cm}{$2\times$} & Conv     & 512 & $1\times1$ / 1 & $14\times14$ \\
        & Conv     & 1024 & $3\times3$ / 1 & $14\times14$ \\
        \bottomrule
        & Conv     & 1024 & $3\times3$ / 1 & $14\times14$ \\
        & Conv     & 1024 & $3\times3$ / 2 & $7\times7$ \\
        & Conv     & 1024 & $3\times3$ / 1 & $7\times7$ \\
        & Conv     & 1024 & $3\times3$ / 1 & $7\times7$ \\
        & FC       &      & 4096           & 4096\\
        & Dropout 0.5 &      &                & 4096\\
        & FC       &      &$7\times7\times30$ & $7\times7\times30$ \\
 \hline
\end{tabular}
\end{table}

\subsection{YOLOv1 Training}
The authors pre-trained the first 20 layers of YOLO at a resolution of $224\times224$ using the ImageNet dataset \cite{russakovsky2015imagenet}. Then, they added the last four layers with randomly initialized weights and fine-tuned the model with the PASCAL VOC 2007, and  VOC 2012 datasets \cite{everingham2010pascal} at a resolution of $448\times448$ to increase the details for more accurate object detection.  

For augmentations, the authors used random scaling and translations of at most 20\% of the input image size, as well as random exposure and saturation with an upper-end factor of 1.5 in the HSV color space.

\begin{figure}[ht]
  \centering
  \includegraphics[width=14cm]{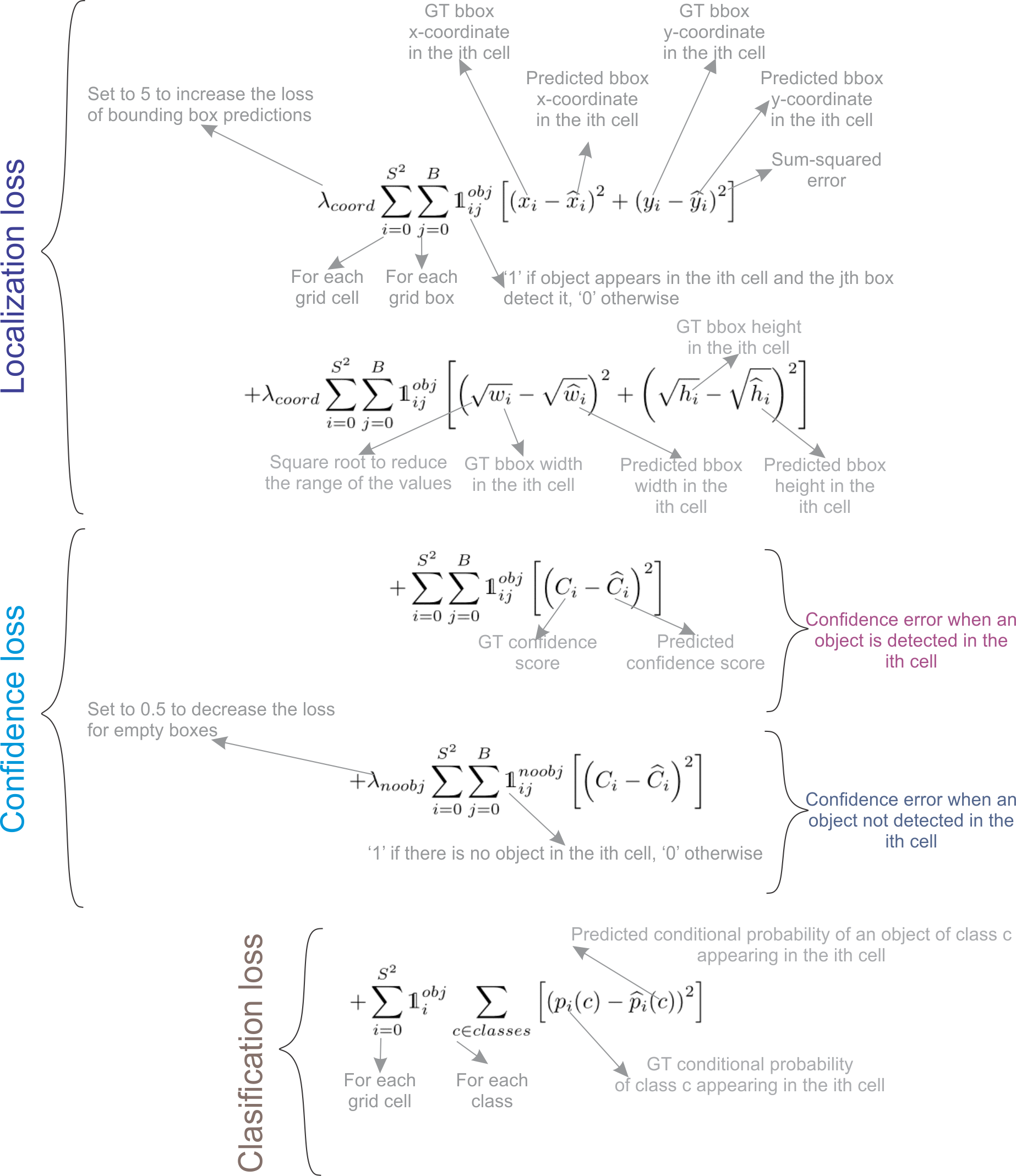}
  \caption{YOLO cost function: includes localization loss for bounding box coordinates, confidence loss for object presence or absence, and classification loss for category prediction accuracy.}
  \label{fig:yolov1_loss}
\end{figure}

YOLOv1 used a loss function composed of multiple sum-squared errors, as shown in Figure \ref{fig:yolov1_loss}. In the loss function, $\lambda_{coord}=5$ is a scale factor that gives more importance to the bounding boxes predictions, and $\lambda_{noobj}=0.5$ is a scale factor that decreases the importance of the boxes that do not contain objects. 

The first two terms of the loss represent the \emph{localization loss}; it computes the error in the predicted bounding boxes locations ($x, y$) and sizes ($w, h$). Note that these errors are only computed in the boxes containing objects (represented by the $\mathds{1}^{obj}_{ij}$), only penalizing if an object is present in that grid cell. The third and fourth loss terms represent the \emph{confidence loss}; the third term measures the confidence error when the object is detected in the box ($\mathds{1}^{obj}_{ij}$) and the fourth term measures the confidence error when the object is not detected in the box ($\mathds{1}^{noobj}_{ij}$). Since most boxes are empty, this loss is weighted down by the $\lambda_{noobj}$ term. The final loss component is the \emph{classification loss} that measures the squared error of the class conditional probabilities for each class only if the object appears in the cell ($\mathds{1}^{obj}_i$).

\subsection{YOLOv1 Strengths and Limitations}
The simple architecture of YOLO, along with its novel full-image one-shot regression, made it much faster than the existing object detectors allowing real-time performance.

However, while YOLO performed faster than any object detector, the localization error was larger compared with state-of-the-art methods such as Fast R-CNN \cite{girshick2015fast}. There were three major causes of this limitation:
\begin{enumerate}
    \item It could only detect at most two objects of the same class in the grid cell, limiting its ability to predict nearby objects.
    \item It struggled to predict objects with aspect ratios not seen in the training data.
    \item It learned from coarse object features due to the down-sampling layers.
\end{enumerate}

\section{YOLOv2: Better, Faster, and Stronger}
\begin{figure}[ht]
  \centering
  \includegraphics[width=10cm]{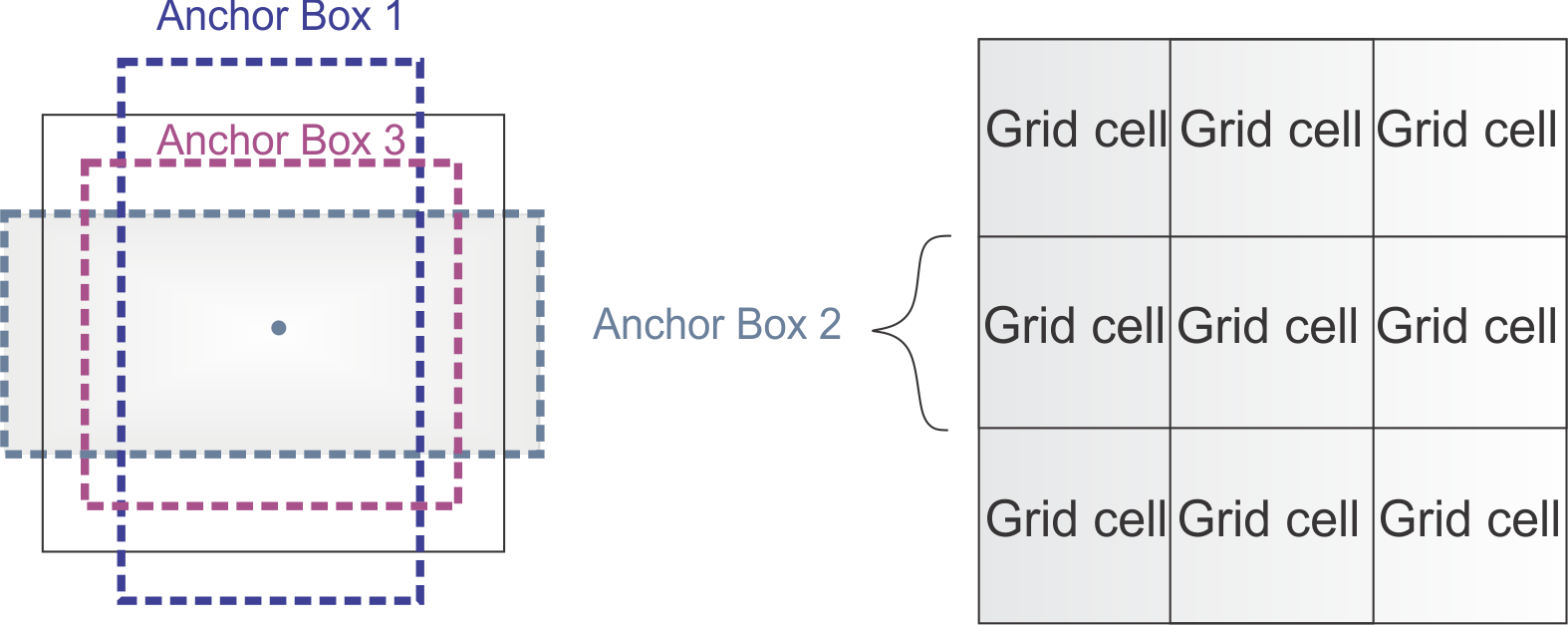}
  \caption{Anchor boxes. YOLOv2 defines multiple anchor boxes for each grid cell.}
  \label{fig:anchor_boxes}
\end{figure}

\begin{figure}[ht]
  \centering
  \includegraphics[width=8cm]{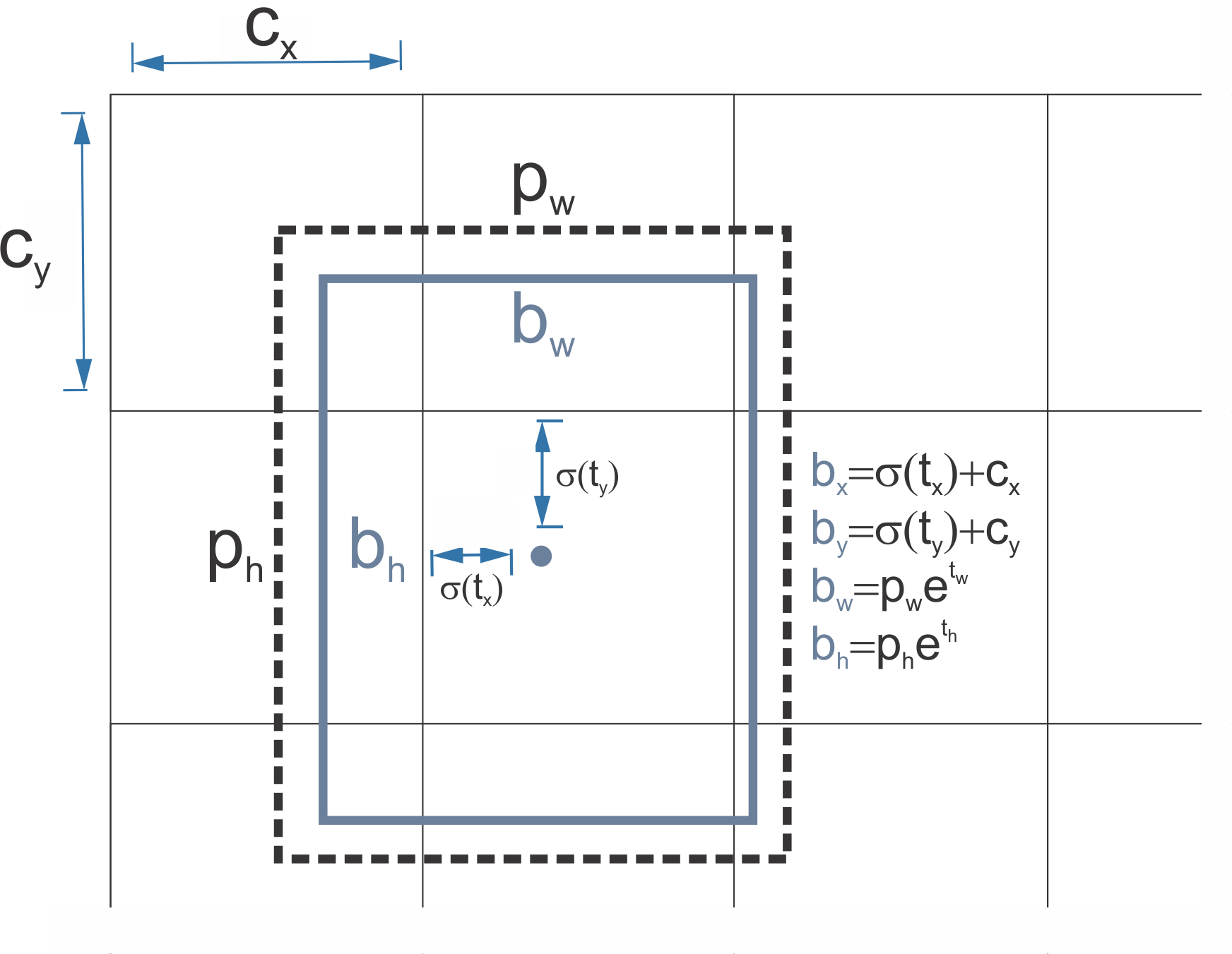}
  \caption{Bounding boxes prediction. The box's center coordinates are obtained with the predicted $t_x$, $t_y$ values passing through a sigmoid function and offset by the location of the grid cell $c_x$, $c_y$. The width and height of the final box use the prior width $p_w$ and height $p_h$ scaled by $e^{t_w}$ and $e^{t_h}$ respectively, where $t_w$ and $t_h$ are predicted by YOLOv2.}
  \label{fig:bbox_with_prior_and_location}
\end{figure}

YOLOv2 was published in CVPR 2017 \cite{redmon2017yolo9000} by Joseph Redmon and Ali Farhadi. It included several improvements over the original YOLO, to make it better, keeping the same speed and also stronger ---capable of detecting 9000 categories!---. The improvements were the following:

\begin{enumerate}
    \item \textbf{Batch normalization} on all convolutional layers improved convergence and acts as a regularizer to reduce overfitting. 
    \item \textbf{High-resolution classifier}. Like YOLOv1, they pre-trained the model with ImageNet at $224\times224$. However, this time, they finetuned the model for ten epochs on ImageNet with a resolution of $448\times448$, improving the network performance on higher resolution input.
    \item \textbf{Fully convolutional}. They removed the dense layers and used a fully convolutional architecture.
    \item \textbf{Use anchor boxes to predict bounding boxes}. They use a set of \emph{prior boxes} or \emph{anchor boxes}, which are boxes with predefined shapes used to match prototypical shapes of objects as shown in Figure \ref{fig:anchor_boxes}. Multiple anchor boxes are defined for each grid cell, and the system predicts the coordinates and the class for every anchor box. The size of the network output is proportional to the number of anchor boxes per grid cell. 
    \item \textbf{Dimension Clusters}. Picking good prior boxes helps the network learn to predict more accurate bounding boxes. The authors ran k-means clustering on the training bounding boxes to find good priors. They selected five prior boxes providing a good tradeoff between recall and model complexity.
    \item \textbf{Direct location prediction}. Unlike other methods that predicted offsets \cite{ren2015faster}, YOLOv2 followed the same philosophy and predicted location coordinates relative to the grid cell. The network predicts five bounding boxes for each cell, each with five values $t_x$, $t_y$, $t_w$, $t_h$, and $t_o$, where $t_o$ is equivalent to $Pc$ from YOLOv1 and the final bounding box coordinates are obtained as shown in Figure \ref{fig:bbox_with_prior_and_location}.
    \item \textbf{Finner-grained features}. YOLOv2, compared with YOLOv1, removed one pooling layer to obtain an output feature map or grid of $13\times13$ for input images of $416\times416$. YOLOv2 also uses a passthrough layer that takes the $26\times 26 \times 512$ feature map and reorganizes it by stacking adjacent features into different channels instead of losing them via a spatial subsampling. This generates $13\times13\times2048$ feature maps concatenated in the channel dimension with the lower resolution $13\times13\times1024$ maps to obtain $13\times13\times3072$ feature maps. See Table \ref{tab:yolov2} for the architectural details.
    \item \textbf{Multi-scale training}. Since YOLOv2 does not use fully connected layers, the inputs can be different sizes. To make YOLOv2 robust to different input sizes, the authors trained the model randomly, changing the input size ---from $320\times320$ up to $608\times608$--- every ten batches. 
\end{enumerate}

With all these improvements, YOLOv2 achieved an average precision (AP) of 78.6\% on the PASCAL VOC2007 dataset compared to the 63.4\% obtained by YOLOv1.

\subsection{YOLOv2 Architecture}
The backbone architecture used by YOLOv2 is called \emph{Darknet-19}, containing 19 convolutional layers and five max-pooling layers. Similar to the architecture of YOLOv1, it is inspired in the Network in Network \cite{lin2013network} using $1\times1$ convolutions between the $3\times3$ to reduce the number of parameters. In addition, as mentioned above, they use batch normalization to regularize and help convergence. 

Table \ref{tab:yolov2} shows the entire Darknet-19 backbone with the object detection head. YOLOv2 predicts five bounding boxes, each with five values and 20 classes when using the PASCAL VOC dataset. 

The object classification head replaces the last four convolutional layers with a single convolutional layer with 1000 filters, followed by a global average pooling layer and a Softmax.

\begin{table}
  \centering
  \caption{YOLOv2 Architecture. Darknet-19 backbone (layers 1 to 23) plus the detection head composed of the last four convolutional layers and the passthrough layer that reorganizes the features of the 17\textsuperscript{th} output of $26 \times 26 \times 512$ into  $13\times 13 \times 2048$ followed by concatenation with the 25\textsuperscript{th} layer. The final convolution generates a grid of $13\times13$ with 125 channels to accommodate 25 predictions (5 coordinates + 20 classes) for five bounding boxes.}
  \label{tab:yolov2}
  \begin{tabular}{lllll}
   \hline
    Num & Type & Filters & Size/Stride & Output\\ 
    \hline
    1   &  Conv/BN     & 32  & $3\times3$ / 1 & $416\times416\times32$\\
    2   &  Max Pool &     & $2\times2$ / 2 & $208\times208\times32$\\
    3   &  Conv/BN    & 64 & $3\times3$ / 1 & $208\times208\times64$\\
    4   &  Max Pool &     & $2\times2$ / 2 & $104\times104\times64$\\
    5   &  Conv/BN     & 128 & $3\times3$ / 1 & $104\times104\times128$ \\   
    6   &  Conv/BN     & 64  & $1\times1$ / 1 & $104\times104\times64$ \\
    7   &  Conv/BN  & 128 & $3\times3$ / 1 & $104\times104\times128$\\
    8   &  Max Pool &     & $2\times2$ / 2 & $52\times52\times128$\\
    9   &  Conv/BN  & 256 & $3\times3$ / 1 & $52\times52\times256$\\
    10   &  Conv/BN  & 128 & $1\times1$ / 1 & $52\times52\times128$\\
    11  &  Conv/BN  & 256 & $3\times3$ / 1 & $52\times52\times256$\\
    12  &  Max Pool &     & $2\times2$ / 2 & $52\times52\times256$\\
    13  &  Conv/BN  & 512 & $3\times3$ / 1 & $26 \times 26 \times 512$\\
    14  &  Conv/BN  & 256 & $1\times1$ / 1 & $26\times26\times256$\\
    15  &  Conv/BN  & 512 & $3\times3$ / 1 & $26\times26\times512$\\
    16  &  Conv/BN  & 256 & $1\times1$ / 1 & $26\times26\times256$\\
    17  &  Conv/BN  & 512 & $3\times3$ / 1 & $26\times26\times512$\\
    18  &  Max Pool &     & $2\times2$ / 2 & $13\times13\times512$\\
    19  &  Conv/BN  & 1024 & $3\times3$ / 1 & $13\times13\times1024$\\
    20  &  Conv/BN  & 512 & $1\times1$ / 1 & $13\times13\times512$\\
    21  &  Conv/BN  & 1024 & $3\times3$ / 1 & $13\times13\times1024$\\
    22  &  Conv/BN  & 512 & $1\times1$ / 1 & $13\times13\times512$\\
    23  &  Conv/BN  & 1024 & $3\times3$ / 1 & $13\times13\times1024$\\
 \hline
    24  &  Conv/BN  & 1024 & $3\times3$ / 1 & $13\times13\times1024$\\
    25  &  Conv/BN  & 1024 & $3\times3$ / 1 & $13\times13\times1024$\\
    26  &  Reorg layer 17        &  &       & $13\times 13 \times 2048$\\
    27  &  Concat 25 and 26      & &        & $13\times13\times3072$\\
    28  &  Conv/BN  & 1024 & $3\times3$ / 1 & $13\times13\times1024$\\
    29  &  Conv  & 125 & $1\times1$ / 1 & $13\times13\times125$\\
\hline
\end{tabular}
\end{table}

\subsection{YOLO9000 is a stronger YOLOv2}
The authors introduced a method for training joint classification and detection in the same paper. It used the detection labeled data from COCO \cite{lin2014microsoft} to learn bounding box coordinates and classification data from ImageNet to increase the number of categories it can detect.   
During training, they combined both datasets such that when a detection training image is used, it backpropagates the detection network, and when a classification training image is used, it backpropagates the classification part of the architecture. The result is a YOLO model capable of detecting more than 9000 categories hence the name YOLO9000.

\section{YOLOv3}
YOLOv3 \cite{redmon2018yolov3} was published in ArXiv in 2018 by Joseph Redmon and Ali Farhadi. It included significant changes and a bigger architecture to be on par with the state-of-the-art while keeping real-time performance. In the following, we described the changes with respect to YOLOv2.
\begin{enumerate}
    \item \textbf{Bounding box prediction}. Like YOLOv2, the network predicts four coordinates for each bounding box $t_x$, $t_y$, $t_w$, and $t_h$; however, this time, YOLOv3 predicts an \emph{objectness score} for each bounding box using logistic regression. This score is 1 for the anchor box with the highest overlap with the ground truth and 0 for the rest anchor boxes. YOLOv3, as opposed to Faster R-CNN \cite{ren2015faster}, assigns only one anchor box to each ground truth object. Also, if no anchor box is assigned to an object, it only incurs in classification loss but not localization loss or confidence loss.
    \item \textbf{Class Prediction}. Instead of using a softmax for the classification, they used binary cross-entropy to train independent logistic classifiers and pose the problem as a multilabel classification. This change allows assigning multiple labels to the same box, which may occur on some complex datasets \cite{krasin2017openimages} with overlapping labels. For example, the same object can be a \emph{Person} and a \emph{Man}.
    \item \textbf{New backbone}. YOLOv3 features a larger feature extractor composed of 53 convolutional layers with residual connections. Section \ref{subsec:yolov3_arch} describes the architecture in more detail.
    \item \textbf{Spatial pyramid pooling (SPP)} Although not mentioned in the paper, the authors also added to the backbone a modified SPP block \cite{he2015spatial} that concatenates multiple max pooling outputs without subsampling (stride = 1), each with different kernel sizes $k \times k$ where $k={1, 5,9,13}$ allowing a larger receptive field. This version is called YOLOv3-spp and was the best-performed version improving the AP\textsubscript{50} by 2.7\%. 
    \item \textbf{Multi-scale Predictions}. Similar to Feature Pyramid Networks \cite{lin2017feature}, YOLOv3 predicts three boxes at three different scales. Section \ref{subsec:yolov3_multiscale} describes the multi-scale prediction mechanism with more details.
    \item \textbf{Bounding box priors}. Like YOLOv2, the authors also use k-means to determine the bounding box priors of anchor boxes. The difference is that in YOLOv2, they used a total of five prior boxes per cell, and in YOLOv3, they used three prior boxes for three different scales.
\end{enumerate}

\subsection{YOLOv3 Architecture}
\label{subsec:yolov3_arch}
\begin{figure}[ht]
  \centering
  \includegraphics[width=\linewidth]{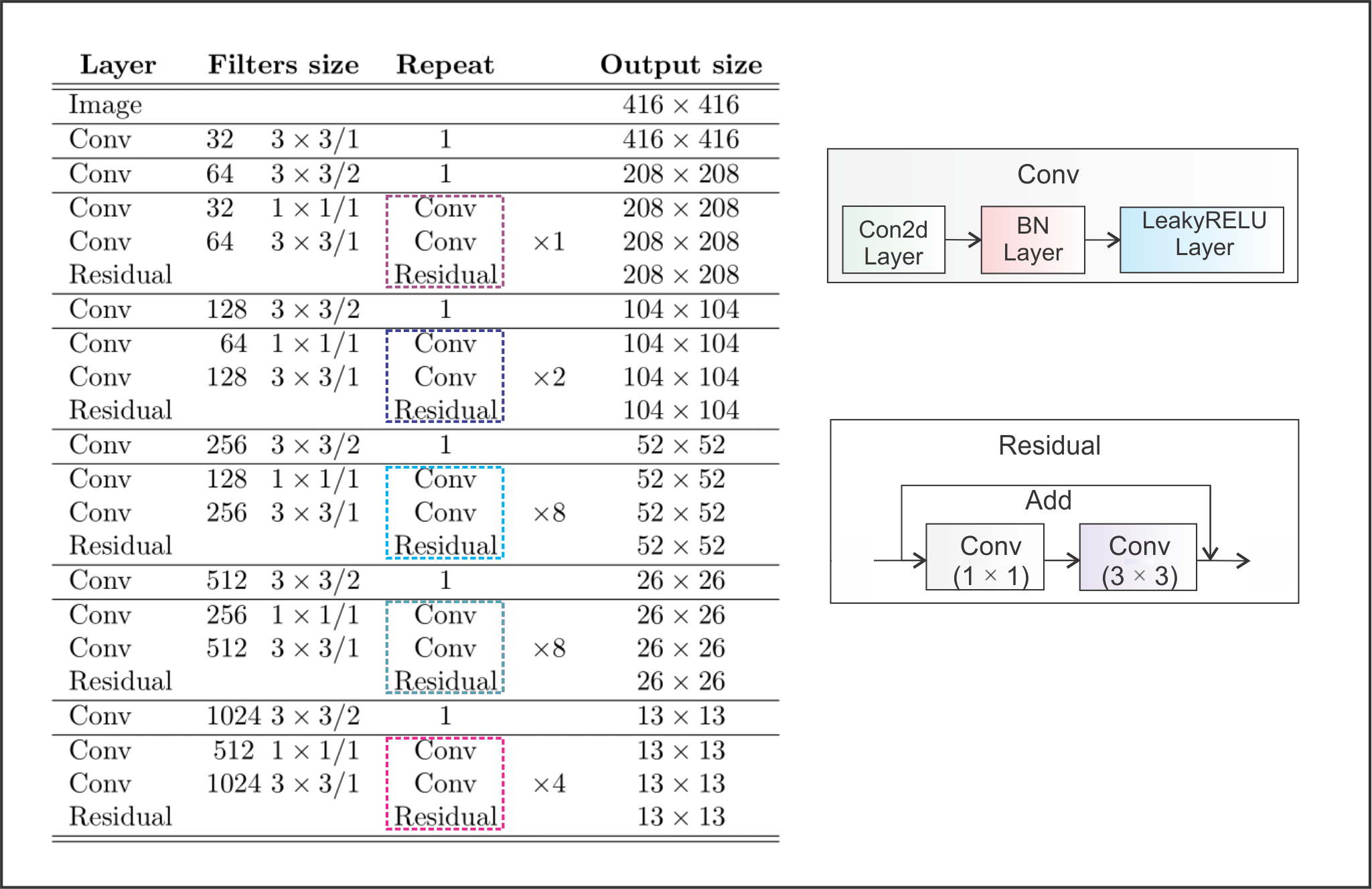}
  \caption{YOLOv3 Darknet-53 backbone. The architecture of YOLOv3 is composed of 53 convolutional layers, each with batch normalization and Leaky ReLU activation. Also, residual connections connect the input of the $1\times1$ convolutions across the whole network with the output of the $3\times3$ convolutions. The architecture shown here consists of only the backbone; it does not include the detection head composed of multi-scale predictions.}
  \label{fig:darknet-53}
\end{figure}

The architecture backbone presented in YOLOv3 is called Darknet-53. It replaced all max-pooling layers with strided convolutions and added residual connections. In total, it contains 53 convolutional layers. Figure \ref{fig:darknet-53} shows the architecture details.

The Darknet-53 backbone obtains Top-1 and Top-5 accuracies comparable with ResNet-152 but almost $2\times$ faster.

\subsection{YOLOv3 Multi-Scale Predictions}
\label{subsec:yolov3_multiscale}
Besides a larger architecture, an essential feature of YOLOv3 is the multi-scale predictions, i.e., predictions at multiple grid sizes. This helped to obtain finer detailed boxes and significantly improved the prediction of small objects, which was one of the main weaknesses of the previous versions of YOLO.  

\begin{figure}[ht]
  \centering
  \includegraphics[width=\linewidth]{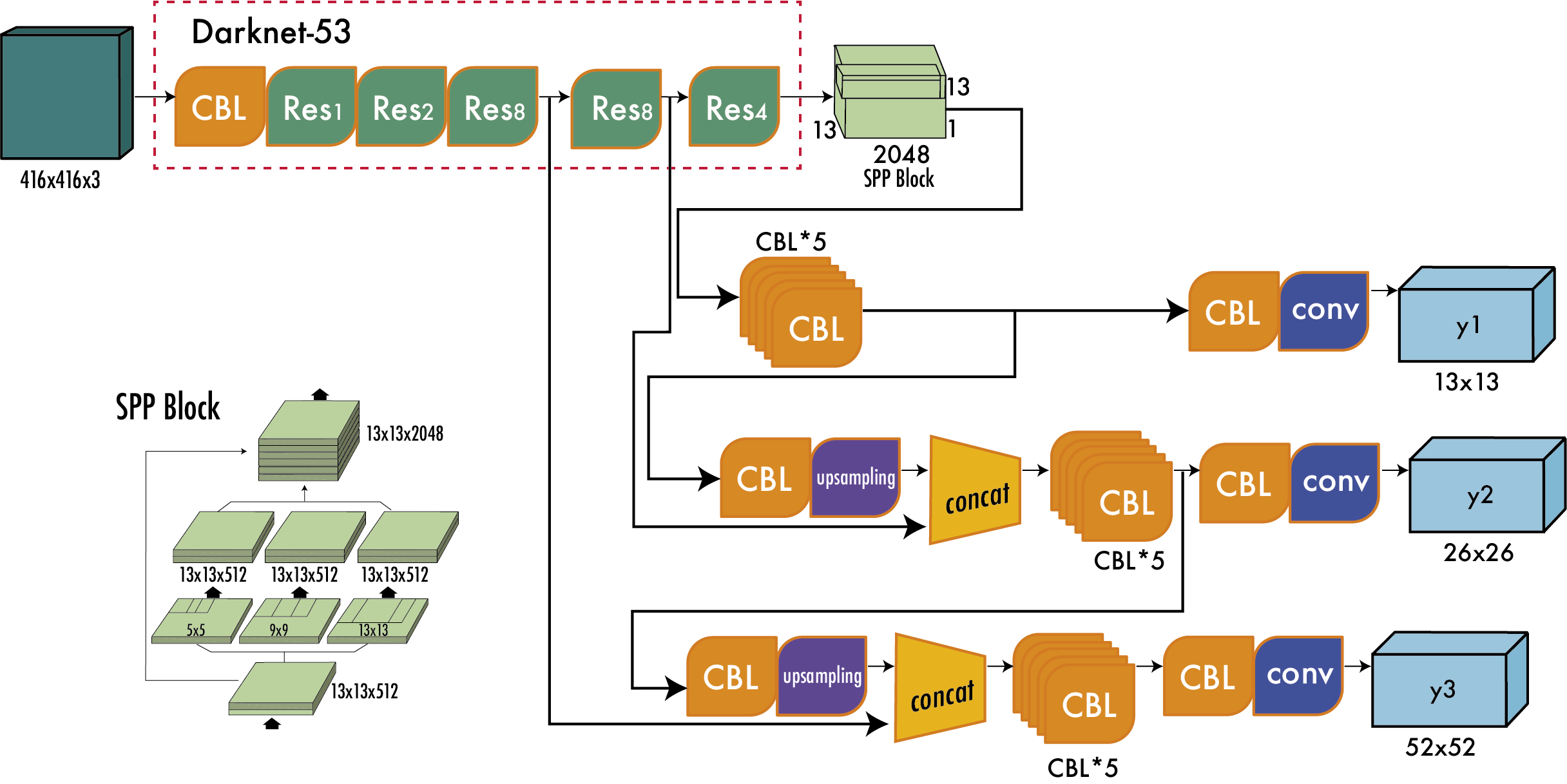}
  \caption{YOLOv3 Multi-scale detection architecture. The output of the Darknet-53 backbone is branched to three different outputs marked as \textbf{y1}, \textbf{y2}, and \textbf{y3}, each of increased resolution. The final predicted boxes are filtered using Non-maximum suppression. The CBL (Convolution-BatchNorm-Leaky ReLU) blocks comprise one convolution layer with batch normalization and leaky ReLU. The Res blocks comprise one CBL followed by two CBL structures with a residual connection, as shown in Figure \ref{fig:darknet-53}.}
  \label{fig:yolov3-multiscale}
\end{figure}

The multi-scale detection architecture shown in Figure \ref{fig:yolov3-multiscale} works as follows: the first output marked as \textbf{y1} is equivalent to the YOLOv2 output, where a $13\times13$ grid defines the output. The second output \textbf{y2} is composed by concatenating the output after the ($Res\times4$) of Darknet-53 with the output after (the $Res\times8$). The feature maps have different sizes, i.e., $13\times13$ and $26\times26$, so there is an upsampling operation before the concatenation. Finally, using an upsampling operation, the third output \textbf{y3} concatenates the $26\times26$ feature maps with the $52\times52$ feature maps.  

For the COCO dataset with 80 categories, each scale provides an output tensor with a shape of $N \times N \times [3 \times (4 + 1 + 80)]$ where $N \times N$ is the size of the feature map (or grid cell), the 3 indicates the boxes per cell and the $4 + 1$ include the four coordinates and the objectness score.

\subsection{YOLOv3 Results}
When YOLOv3 was released, the benchmark for object detection had changed from PASCAL VOC to Microsoft COCO \cite{lin2014microsoft}. Therefore, from here on, all the YOLOs are evaluated in the MS COCO dataset.  
YOLOv3-spp achieved an average precision AP of 36.2\% and AP\textsubscript{50} of 60.6\% at 20 FPS, achieving state-of-the-art at the time and $2\times$ faster.

\section{Backbone, Neck, and Head}
\begin{figure}[ht]
  \centering
  \includegraphics[width=13cm]{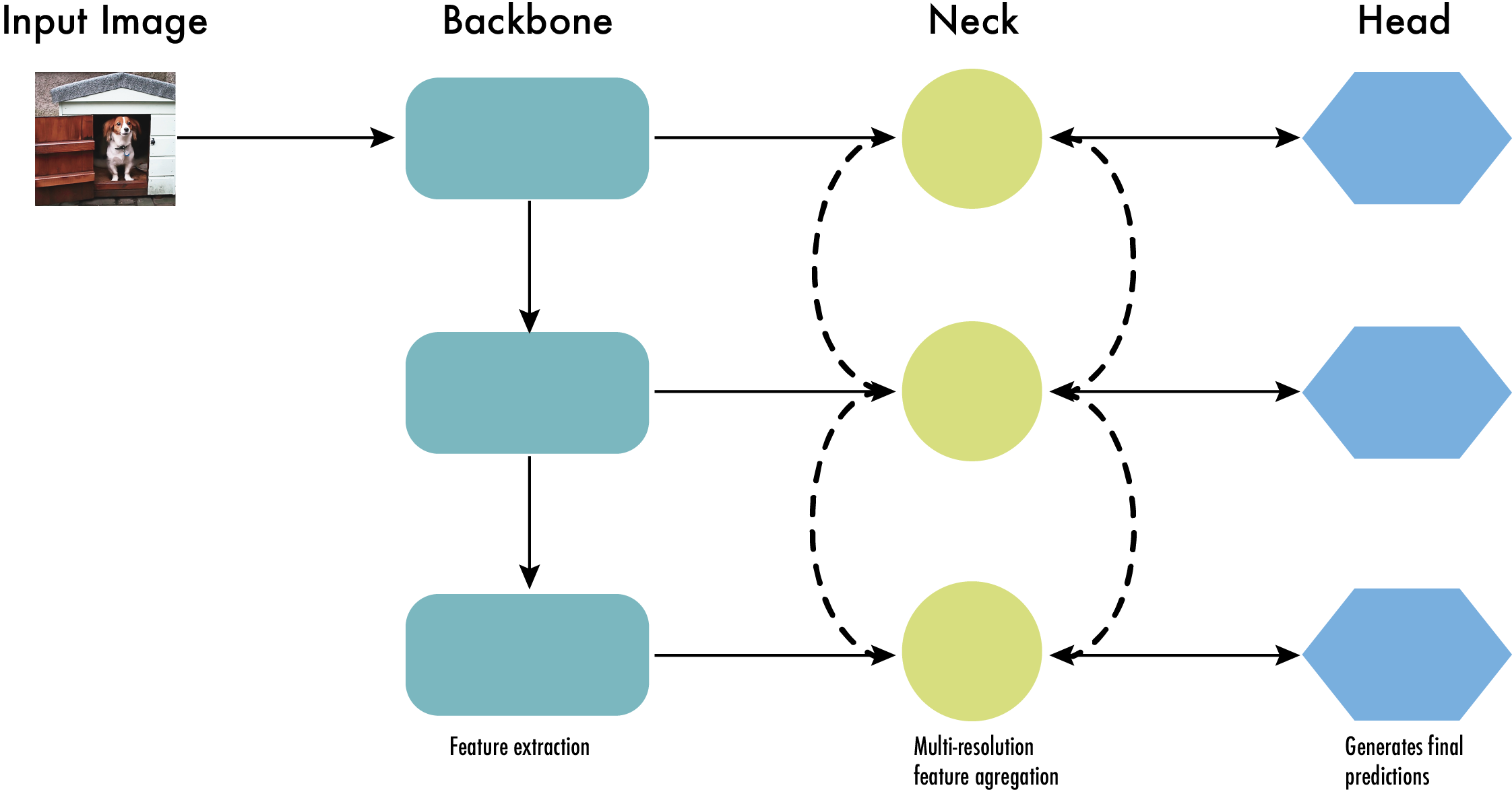}
  \caption{The architecture of modern object detectors can be described as the backbone, the neck, and the head. The backbone, usually a convolutional neural network (CNN), extracts vital features from the image at different scales. The neck refines these features, enhancing spatial and semantic information. Lastly, the head uses these refined features to make object detection predictions.}
  \label{fig:backbone-neck-head}
\end{figure}

At this time, the architecture of object detectors started to be described in three parts: the backbone, the neck, and the head. Figure \ref{fig:backbone-neck-head} shows a high-level backbone, neck, and head diagram.

The backbone is responsible for extracting useful features from the input image. It is typically a convolutional neural network (CNN) trained on a large-scale image classification task, such as ImageNet. The backbone captures hierarchical features at different scales, with lower-level features (e.g., edges and textures) extracted in the earlier layers and higher-level features (e.g., object parts and semantic information) extracted in the deeper layers.

The neck is an intermediate component that connects the backbone to the head. It aggregates and refines the features extracted by the backbone, often focusing on enhancing the spatial and semantic information across different scales. The neck may include additional convolutional layers, feature pyramid networks (FPN) \cite{lin2017feature}, or other mechanisms to improve the representation of the features. 

The head is the final component of an object detector; it is responsible for making predictions based on the features provided by the backbone and neck. It typically consists of one or more task-specific subnetworks that perform classification, localization, and, more recently, instance segmentation and pose estimation. The head processes the features the neck provides, generating predictions for each object candidate. In the end, a post-processing step, such as non-maximum suppression (NMS), filters out overlapping predictions and retains only the most confident detections.

In the rest of the YOLO models, we will describe the architectures using the backbone, neck, and head.

\section{YOLOv4}
Two years passed, and there was no new version of YOLO. It was until April 2020 that Alexey Bochkovskiy, Chien-Yao Wang, and Hong-Yuan Mark Liao released in ArXiv the paper for YOLOv4 \cite{bochkovskiy2020yolov4}. At first, it felt odd that different authors presented a new "official" version of YOLO; however, YOLOv4 kept the same YOLO philosophy ---real-time, open source, single shot, and darknet framework--- and the improvements were so satisfactory that the community rapidly embrace this version as the official YOLOv4.  

YOLOv4 tried to find the optimal balance by experimenting with many changes categorized as \emph{bag-of-freebies} and \emph{bag-of-specials}. Bag-of-freebies are methods that only change the training strategy and increase training cost but do not increase the inference time, the most common being data augmentation. On the other hand, bag-of-specials are methods that slightly increase the inference cost but significantly improve accuracy. Examples of these methods are those for enlarging the receptive field \cite{he2015spatial, chen2017deeplab, liu2018receptive}, combining features \cite{he2016deep, lin2017feature, hariharan2015hypercolumns, zhao2019m2det}, and post-processing \cite{he2015delving, maas2013rectifier, misra2019mish, bodla2017soft} among others.

\begin{figure}[ht]
  \centering
  \includegraphics[width=\linewidth]{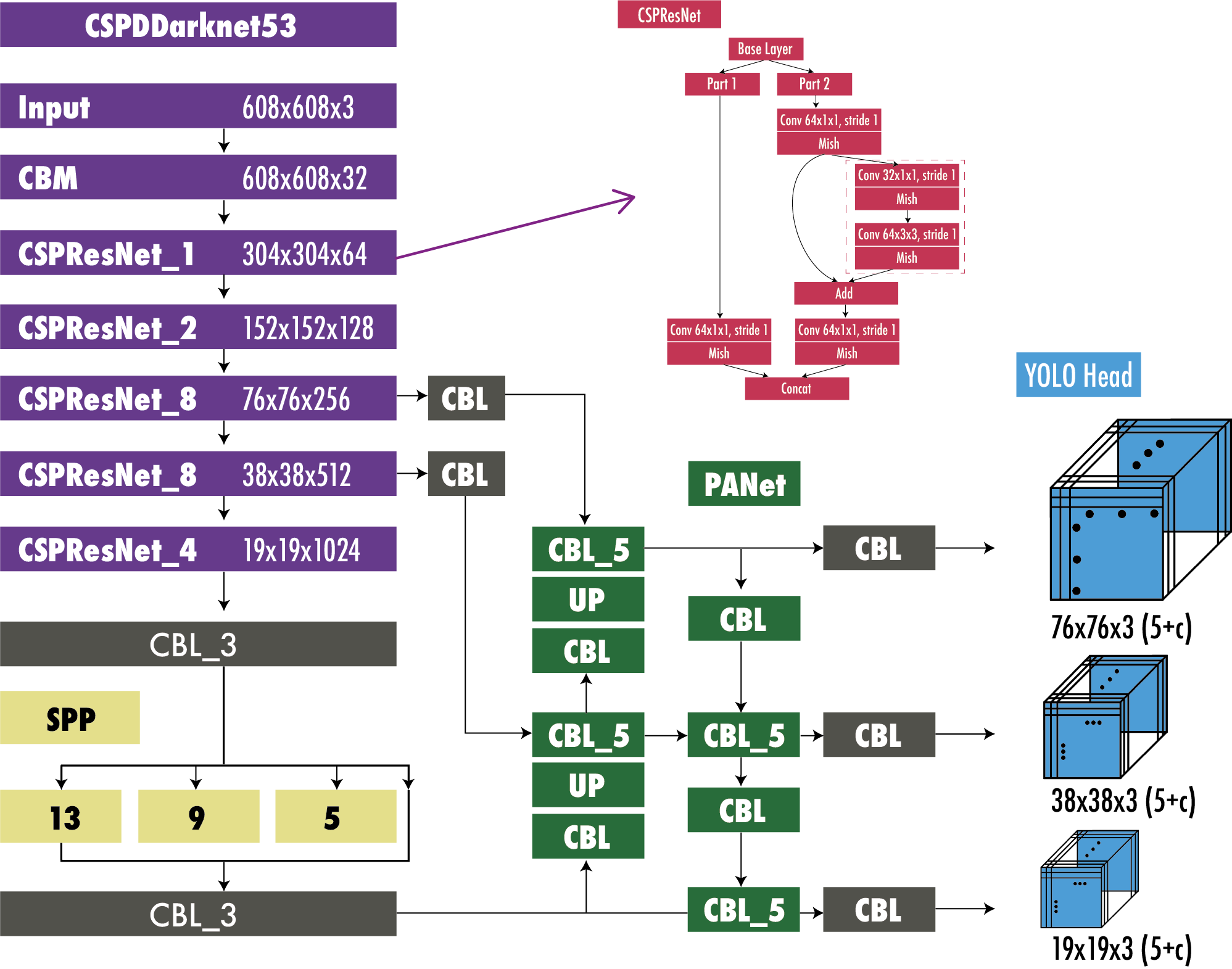}
  \caption{YOLOv4 Architecture for object detection. The modules in the diagram are \textbf{CMB}: Convolution + Batch Normalization + Mish activation, \textbf{CBL}: Convolution + Batch Normalization + Leaky ReLU, \textbf{UP}: upsampling, \textbf{SPP}: Spatial Pyramid Pooling, and \textbf{PANet}: Path Aggregation Network. Diagram inspired by \cite{wang2021real}.}
  \label{fig:yolov4-architecture}
\end{figure}

We summarize the main changes of YOLOv4 in the following points:

\begin{itemize}
    \item \textbf{An Enhanced Architecture with Bag-of-Specials (BoS) Integration}. The authors tried multiple architectures for the backbone, such as ResNeXt50 \cite{xie2017aggregated}, EfficientNet-B3 \cite{tan2019efficientnet}, and Darknet-53. The best-performing architecture was a modification of Darknet-53 with cross-stage partial connections (CSPNet) \cite{wang2020cspnet}, and Mish activation function \cite{misra2019mish} as the backbone (see Figure \ref{fig:yolov4-architecture}. For the neck, they used the modified version of spatial pyramid pooling (SPP) \cite{he2015spatial} from YOLOv3-spp and multi-scale predictions as in YOLOv3, but with a modified version of path aggregation network (PANet) \cite{liu2018path} instead of FPN as well as a modified spatial attention module (SAM) \cite{woo2018cbam}. Finally, for the detection head, they use anchors as in YOLOv3. Therefore, the model was called \emph{CSPDarknet53-PANet-SPP}. 
    The cross-stage partial connections (CSP) added to the Darknet-53 help reduce the computation of the model while keeping the same accuracy. The SPP block, as in YOLOv3-spp increases the receptive field without affecting the inference speed. The modified version of PANet concatenates the features instead of adding them as in the original PANet paper.
    \item \textbf{Integrating bag-of-freebies (BoF) for an Advanced Training Approach}. Apart from the regular augmentations such as random brightness, contrast, scaling, cropping, flipping, and rotation, the authors implemented mosaic augmentation that combines four images into a single one allowing the detection of objects outside their usual context and also reducing the need for a large mini-batch size for batch normalization. For regularization, they used DropBlock \cite{ghiasi2018dropblock} that works as a replacement of Dropout \cite{srivastava2014dropout} but for convolutional neural networks as well as class label smoothing \cite{szegedy2016rethinking, islam2017label}.
    For the detector, they added CIoU loss \cite{zheng2020distance} and Cross mini-bath normalization (CmBN) for collecting statistics from the entire batch instead of from single mini-batches as in regular batch normalization \cite{ioffe2015batch}.
    \item\textbf{Self-adversarial Training (SAT)}. To make the model more robust to perturbations, an adversarial attack is performed on the input image to create a deception that the ground truth object is not in the image but keeps the original label to detect the correct object. 
    \item \textbf{Hyperparameter Optimization with Genetic Algorithms}. To find the optimal hyperparameters used for training, they use genetic algorithms on the first 10\% of periods, and a cosine annealing scheduler \cite{loshchilov2016sgdr} to alter the learning rate during training. It starts reducing the learning rate slowly, followed by a quick reduction halfway through the training process ending with a slight reduction.
\end{itemize}

Table \ref{tab:yolov4-bof-bos} lists the final selection of BoFs and BoS for the backbone and the detector.
 

\begin{table}
  \centering
  \caption{YOLOv4 final selection of bag-of-freebies (BoF) and bag-of-specials (BoS). BoF are methods that increase performance with no inference cost but longer training times. On the other hand, BoS are methods that slightly increase the inference cost but significantly improve accuracy.}
  \label{tab:yolov4-bof-bos}
  \begin{tabular}{ll}
   \hline
    \textbf{Backbone}  & \textbf{Detector} \\ 
    \hline
    \textbf{Bag-of-Freebies}    &   \textbf{Bag-of-Freebies} \\
    Data augmentation  &     Data augmentation \\
    - Mosaic           &     - Mosaic \\
    - CutMix           &     - Self-Adversarial Training \\
    Regularization     &     CIoU loss \\
    - DropBlock        &     Cross mini-Batch Normalization (CmBN) \\
    Class label smoothing  & Eliminate grid sensitivity \\
                       &     Multiple anchors for a single ground truth \\
                       &     Cosine annealing scheduler \\
                       &     Optimal hyper-parameteres \\
                       &     Random training shapes \\
 \\
    \textbf{Bag-of-Specials}    &   \textbf{Bag-of-Specials} \\
    Mish activation    &     Mish activation \\
    Cross-stage partial connections    &     Spatial pyramid pooling block \\                       
    Multi-input weighted residual connections    &    Spatial attention module (SAM) \\
                       &     Path aggregation network (PAN) \\
                       &     Distance-IoU Non-Maximum Suppression \\                       
\hline
\end{tabular}
\end{table}

Evaluated on MS COCO dataset test-dev 2017, YOLOv4 achieved an AP of 43.5\% and AP\textsubscript{50} of 65.7\% at more than 50 FPS on an NVIDIA V100.

\section{YOLOv5}
\begin{figure}[ht]
  \centering
  \includegraphics[width= \linewidth]{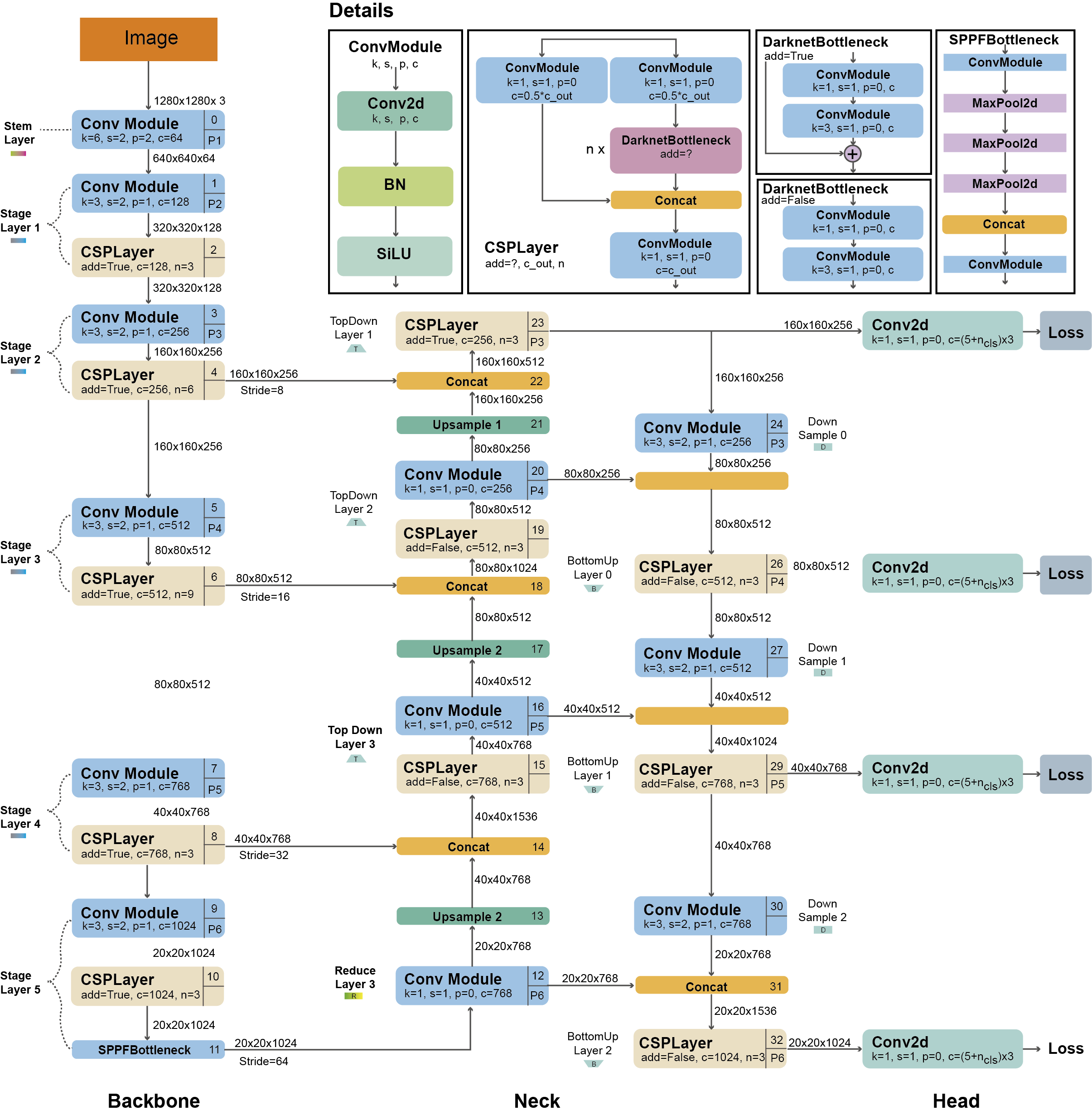}
  \caption{YOLOv5 Architecture. The architecture uses a modified CSPDarknet53 backbone with a Stem, followed by convolutional layers that extract image features. A spatial pyramid pooling fast (SPPF) layer accelerates computation by pooling features into a fixed-size map. Each convolution has batch normalization and SiLU activation. The network's neck uses SPPF and a modified CSP-PAN, while the head resembles YOLOv3. Diagram based in \cite{mmyolo_vyolov5} and \cite{ultralytics_yolov5_arch_summary}.}
  \label{fig:yolov5_arch}
\end{figure}

YOLOv5 \cite{Jocher_YOLOv5_by_Ultralytics_2020} was released a couple of months after YOLOv4 in 2020 by Glen Jocher, founder and CEO of Ultralytics. It uses many improvements described in the YOLOv4 section but developed in Pytorch instead of Darknet. YOLOv5 incorporates an Ultralytics algorithm called AutoAnchor. This pre-training tool checks and adjusts anchor boxes if they are ill-fitted for the dataset and training settings, such as image size. It first applies a k-means function to dataset labels to generate initial conditions for a Genetic Evolution (GE) algorithm. The GE algorithm then evolves these anchors over 1000 generations by default, using CIoU loss \cite{zheng2020distance} and Best Possible Recall as its fitness function.
Figure \ref{fig:yolov5_arch} shows the detailed architecture of YOLOv5.

\subsection{YOLOv5 Architecture}
The backbone is a modified CSPDarknet53 that starts with a Stem, a strided convolution layer with a large window size to reduce memory and computational costs; followed by convolutional layers that extract relevant features from the input image. The SPPF (spatial pyramid pooling fast) layer and the following convolution layers process the features at various scales, while the upsample layers increase the resolution of the feature maps. The SPPF layer aims to speed up the computation of the network by pooling features of different scales into a fixed-size feature map. Each convolution is followed by batch normalization (BN) and SiLU activation \cite{hendrycks2016gaussian}. The neck uses SPPF and a modified CSP-PAN, while the head resembles YOLOv3.

YOLOv5 uses several augmentations such as Mosaic, copy paste \cite{ghiasi2021simple}, random affine, MixUp \cite{zhang2017mixup}, HSV augmentation, random horizontal flip, as well as other augmentations from the albumentations package \cite{albu_info11020125}. It also improves the grid sensitivity to make it more stable to runaway gradients.

YOLOv5 provides five scaled versions: YOLOv5n (nano), YOLOv5s (small), YOLOv5m (medium), YOLOv5l (large), and YOLOv5x (extra large), where the width and depth of the convolution modules vary to suit specific applications and hardware requirements. For instance, YOLOv5n and YOLOv5s are lightweight models targeted for low-resource devices, while YOLOv5x is optimized for high performance, albeit at the expense of speed.

The YOLOv5 released version at the time of this writing is v7.0, including YOLOv5 versions capable of classification and instance segmentation.

YOLOv5 is open source and actively maintained by Ultralytics, with more than 250 contributors and new improvements frequently. YOLOv5 is easy to use, train and deploy. Ultralytics provide a mobile version for iOS and Android and many integrations for labeling, training, and deployment.

Evaluated on MS COCO dataset test-dev 2017, YOLOv5x achieved an AP of 50.7\% with an image size of 640 pixels. Using a batch size of 32, it can achieve a speed of 200 FPS on an NVIDIA V100. Using a larger input size of 1536 pixels and test-time augmentation (TTA), YOLOv5 achieves an AP of 55.8\%.

\section{Scaled-YOLOv4}
One year after YOLOv4, the same authors presented Scaled-YOLOv4 \cite{wang2021scaled} in CVPR 2021. Differently from YOLOv4, Scaled YOLOv4 was developed in Pytorch instead of Darknet. The main novelty was the introduction of scaling-up and scaling-down techniques. Scaling up means producing a model that increases accuracy at the expense of a lower speed; on the other hand, scaling down entails producing a model that increases speed sacrificing accuracy. In addition, scaled-down models need less computing power and can run on embedded systems. 

The scaled-down architecture was called YOLOv4-tiny; it was designed for low-end GPUs and can run at 46 FPS on a Jetson TX2 or 440 FPS on RTX2080Ti, achieving 22\% AP on MS COCO.

The scaled-up model architecture was called YOLOv4-large, which included three different sizes P5, P6, and P7. This architecture was designed for cloud GPU and achieved state-of-the-art performance, surpassing all previous models \cite{tan2020efficientdet,lin2017focal,long2020pp} with 56\% AP on MS COCO.

\section{YOLOR}
YOLOR \cite{wang2021you} was published in ArXiv in May 2021 by the same research team of YOLOv4. It stands for \emph{You Only Learn One Representation}. In this paper, the authors followed a different approach; they developed a multi-task learning approach that aims to create a single model for various tasks (e.g., classification, detection, pose estimation) by learning a general representation and using sub-networks to create task-specific representations. With the insight that the traditional joint learning method often leads to suboptimal feature generation, YOLOR aims to overcome this by encoding the implicit knowledge of neural networks to be applied to multiple tasks, similar to how humans use past experiences to approach new problems. The results showed that introducing implicit knowledge into the neural network benefits all the tasks. 

Evaluated on MS COCO dataset test-dev 2017, YOLOR achieved a AP of 55.4\% and AP\textsubscript{50} of 73.3\% at 30 FPS on an NVIDIA V100.

\section{YOLOX}
\begin{figure}[ht]
  \centering
  \includegraphics[width=10cm]{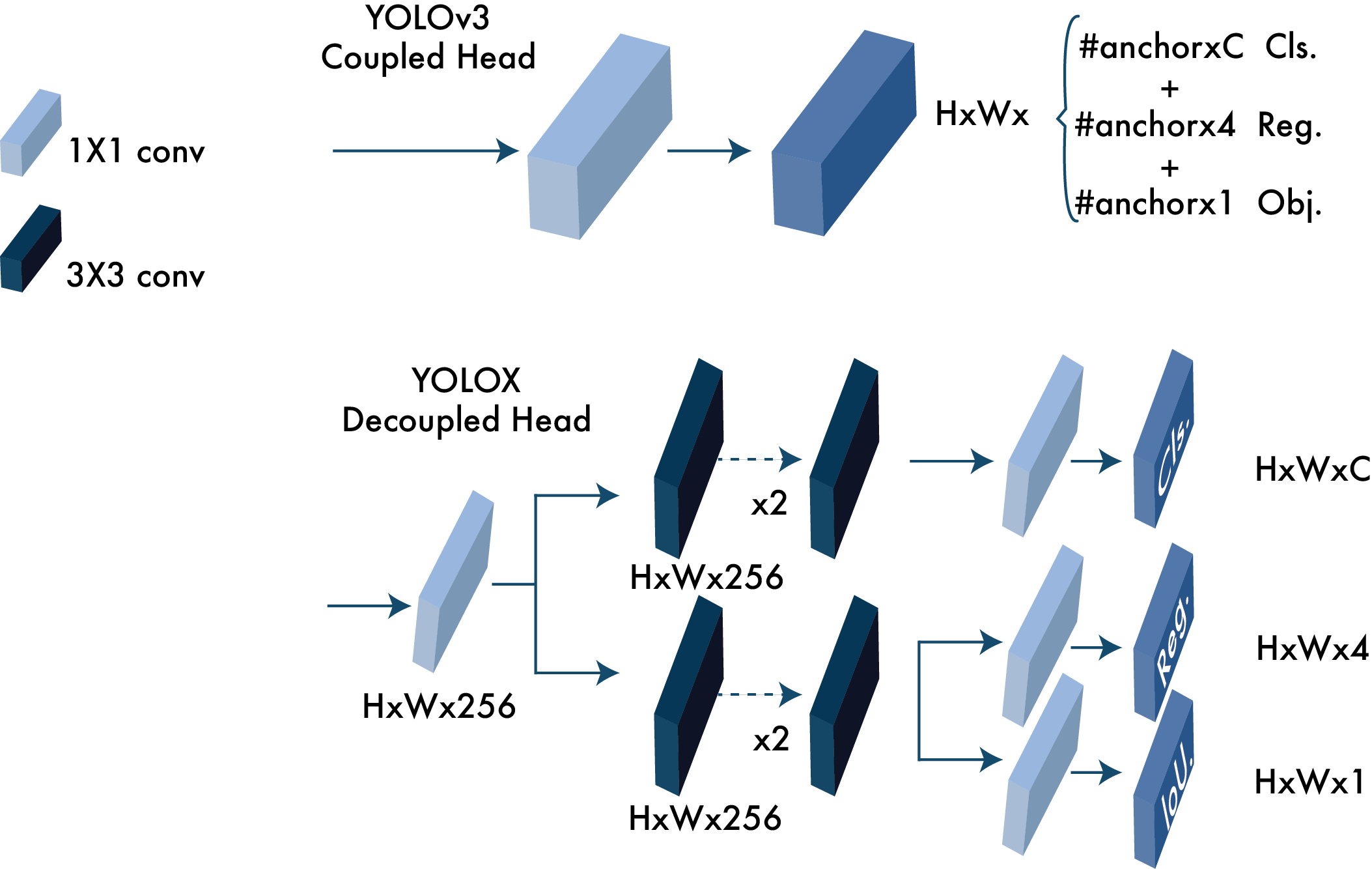}
  \caption{Difference between YOLOv3 head and YOLOX decoupled head. For each level of the FPN,  they used a $1\times1$ convolution layer to reduce the feature channel to 256 and then added two parallel branches with two $3\times3$ convolution layers each for the class confidence (classification) and localization (regression) tasks. The IoU branch is added to the regression head.}
  \label{fig:yolox-decoupled-head}
\end{figure}

YOLOX \cite{ge2021yolox} was published in ArXiv in July 2021 by  Megvii Technology. Developed in Pytorch and using YOLOV3 from Ultralytics as starting point, it has five principal changes: an anchor-free architecture, multiple positives, a decoupled head, advanced label assignment, and strong augmentations. It achieved state-of-the-art results in 2021 with an optimal balance between speed and accuracy with 50.1\% AP at 68.9\% FPS on Tesla V100.  
In the following, we describe the five main changes of YOLOX with respect to YOLOv3:
\begin{enumerate}
    \item \textbf{Anchor-free}. Since YOLOv2, all subsequent YOLO versions were anchor-based detectors. YOLOX, inspired by anchor-free state-of-the-art object detectors such as CornerNet \cite{law2018cornernet}, CenterNet \cite{duan2019centernet}, and FCOS \cite{tian2019fcos}, returned to an anchor-free architecture simplifying the training and decoding process. The anchor-free increased the AP by 0.9 points concerning the YOLOv3 baseline.
    \item \textbf{Multi positives}. To compensate for the large imbalances the lack of anchors produced, the authors use center sampling \cite{tian2019fcos} where they assigned the center $3\times3$ area as positives. This approach increased AP by 2.1 points.
    \item \textbf{Decoupled head}. In \cite{song2020revisiting,wu2020rethinking}, it was shown that there could be a misalignment between the classification confidence and localization accuracy. Due to this, YOLOX separates these two into two heads (as shown in Fig. \ref{fig:yolox-decoupled-head}), one for classification tasks and the other for regression tasks improving the AP by 1.1 points and speeding up the model convergence.
    \item \textbf{Advanced label assignment}. In \cite{ge2021ota}, it was shown that the ground truth label assignment could have ambiguities when the boxes of multiple objects overlap and formulate the assigning procedure as an Optimal Transport (OT) problem. YOLOX, inspired by this work, proposed a simplified version called simOTA. This change increased AP by 2.3 points.
    \item \textbf{Strong augmentations}. YOLOX uses MixUP \cite{zhang2017mixup} and Mosaic augmentations. The authors found that ImageNet pretraining was no longer beneficial after using these augmentations. The strong augmentations increased AP by 2.4 points.
\end{enumerate}

\section{YOLOv6}
YOLOv6 \cite{li2022yolov6} was published in ArXiv in September 2022 by Meituan Vision AI Department. The network design consists of an efficient backbone with RepVGG or CSPStackRep blocks, a PAN topology neck, and an efficient decoupled head with a hybrid-channel strategy. In addition, the paper introduces enhanced quantization techniques using post-training quantization and channel-wise distillation, resulting in faster and more accurate detectors. Overall, YOLOv6 outperforms previous state-of-the-art models on accuracy and speed metrics, such as YOLOv5, YOLOX, and PP-YOLOE. 

\begin{figure}[ht]
  \centering
  \includegraphics[width=\linewidth]{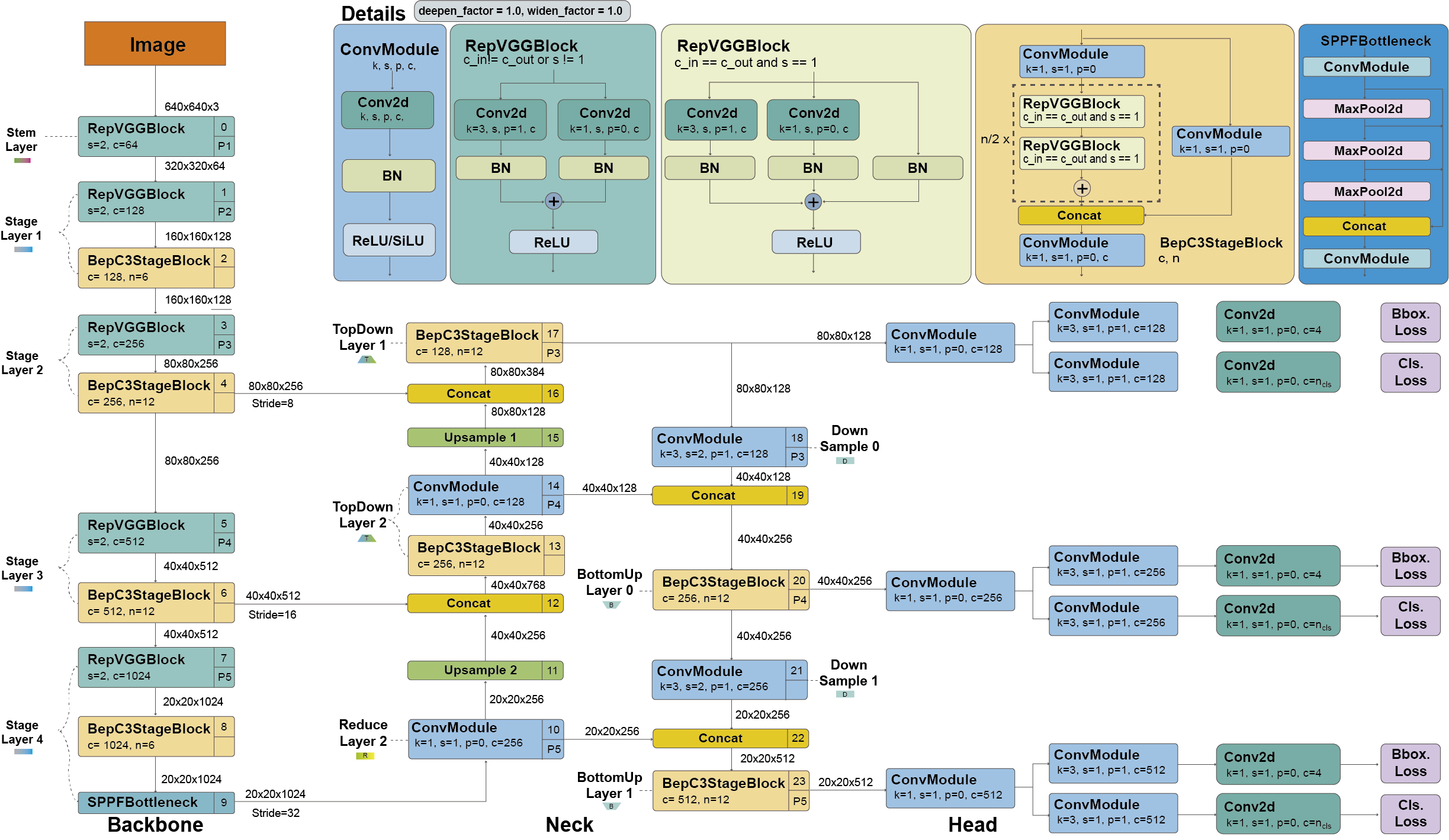}
  \caption{YOLOv6 Architecture. The architecture uses a new backbone with RepVGG blocks \cite{ding2021repvgg}. The spatial pyramid pooling fast (SPPF) and Conv Modules are similar to YOLOv5. However, YOLOv6 uses a decoupled head. Diagram based in \cite{mmyolo_vyolov6}.}
  \label{fig:yolov6_arch}
\end{figure}

Figure \ref{fig:yolov6_arch} shows the detailed architecture of YOLOv6.

The main novelties of this model are summarized below:

\begin{enumerate}
    \item \textbf{A new backbone based on RepVGG} \cite{ding2021repvgg} called EfficientRep that uses higher parallelism than previous YOLO backbones. For the neck, they use PAN \cite{liu2018path} enhanced with RepBlocks \cite{ding2021repvgg} or CSPStackRep\cite{wang2020cspnet} Blocks for the larger models. And following YOLOX, they developed an efficient decoupled head.
    \item \textbf{Label assignment} using the Task alignment learning approach introduced in TOOD \cite{feng2021tood}.
    \item \textbf{New classification and regression losses}. They used a classification VariFocal loss \cite{zhang2021varifocalnet} and an SIoU \cite{gevorgyan2022siou}/GIoU \cite{rezatofighi2019generalized} regression loss.
    \item \textbf{A self-distillation} strategy for the regression and classification tasks.
    \item \textbf{A quantization scheme} for detection using RepOptimizer\cite{ding2022re} and channel-wise distillation \cite{shu2021channel} that helped to achieve a faster detector.
\end{enumerate}

The authors provide eight scaled models, from YOLOv6-N to YOLOv6-L6. Evaluated on MS COCO dataset test-dev 2017, the largest model, achieved an AP of 57.2\%  at around 29 FPS on an NVIDIA Tesla T4.

\section{YOLOv7}

YOLOv7 \cite{wang2022yolov7} was published in ArXiv in July 2022 by the same authors of YOLOv4 and YOLOR. At the time, it surpassed all known object detectors in speed and accuracy in the range of 5 FPS to 160 FPS. Like YOLOv4, it was trained using only the MS COCO dataset without pre-trained backbones. 
YOLOv7 proposed a couple of architecture changes and a series of bag-of-freebies, which increased the accuracy without affecting the inference speed, only the training time. 

\begin{figure}[ht]
  \centering
  \includegraphics[width=\linewidth]{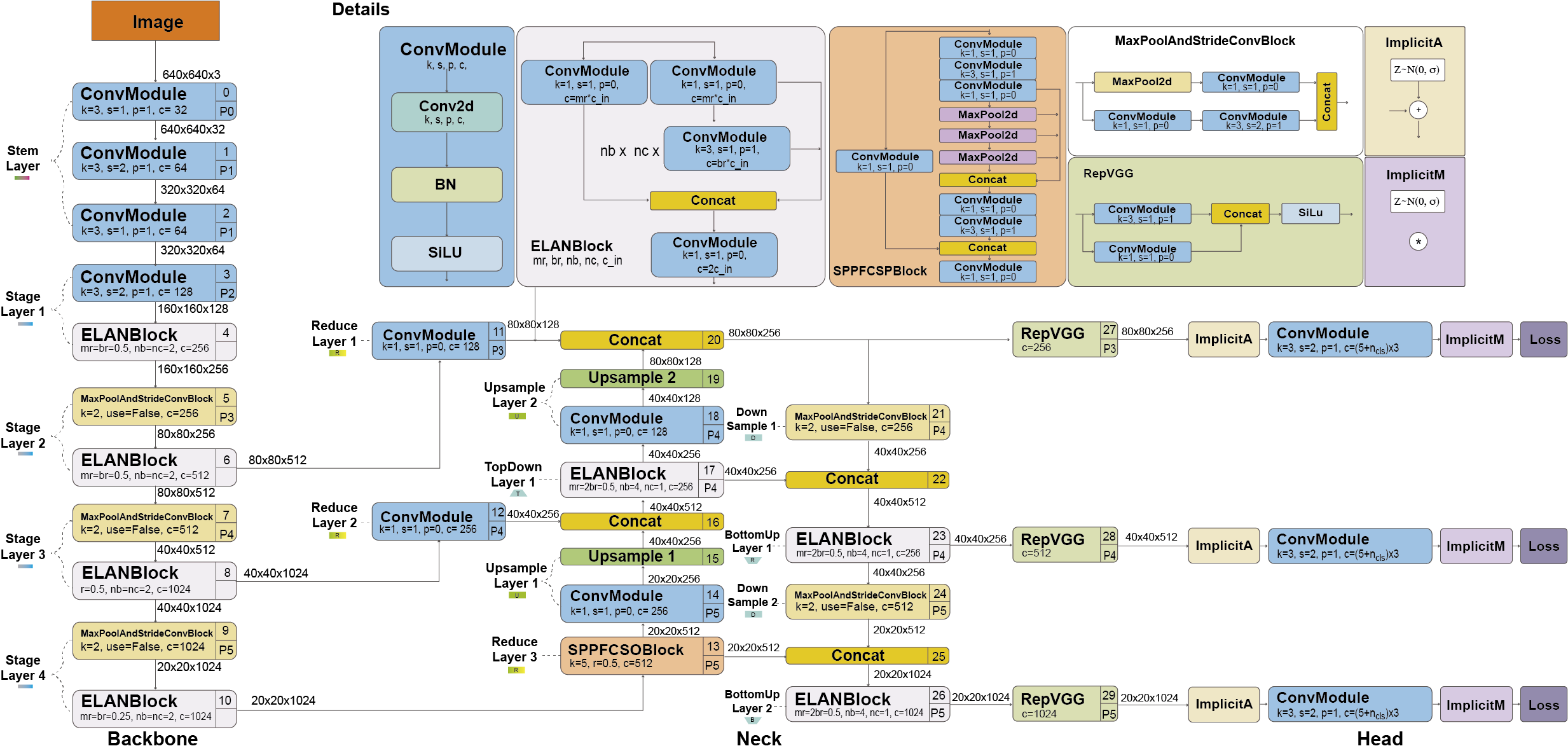}
  \caption{YOLOv7 Architecture. Changes in this architecture include the ELAN blocks that combine features of different groups by shuffling and merging cardinality to enhance the model learning and modified RepVGG without identity connection. Diagram based in \cite{mmyolo_vyolov7}.}
  \label{fig:yolov7_arch}
\end{figure}

Figure \ref{fig:yolov7_arch} shows the detailed architecture of YOLOv7.

The architecture changes of YOLOv7 are:
 \begin{itemize}
     \item \textbf{Extended efficient layer aggregation network (E-ELAN)}. ELAN \cite{wang2022designing} is a strategy that allows a deep model to learn and converge more efficiently by controlling the shortest longest gradient path. YOLOv7 proposed E-ELAN that works for models with unlimited stacked computational blocks. E-ELAN combines the features of different groups by shuffling and merging cardinality to enhance the network's learning without destroying the original gradient path.
     \item \textbf{Model scaling for concatenation-based models}. Scaling generates models of different sizes by adjusting some model attributes. The architecture of YOLOv7 is a concatenation-based architecture in which standard scaling techniques, such as depth scaling, cause a ratio change between the input channel and the output channel of a transition layer which, in turn, leads to a decrease in the hardware usage of the model. YOLOv7 proposed a new strategy for scaling concatenation-based models in which the depth and width of the block are scaled with the same factor to maintain the optimal structure of the model.
\end{itemize}

The bag-of-freebies used in YOLOv7 include:
 \begin{itemize}
     \item \textbf{Planned re-parameterized convolution}. Like YOLOv6, the architecture of YOLOv7 is also inspired by re-parameterized convolutions (RepConv) \cite{ding2021repvgg}. However, they found that the identity connection in RepConv destroys the residual in ResNet \cite{he2016deep} and the concatenation in DenseNet \cite{huang2017densely}. For this reason, they removed the identity connection and called it RepConvN.
     \item \textbf{Coarse label assignment for auxiliary head and fine label assignment for the lead head}. The lead head is responsible for the final output, while the auxiliary head assists with the training. 
     \item \textbf{Batch normalization in conv-bn-activation}. This integrates the mean and variance of batch normalization into the bias and weight of the convolutional layer at the inference stage.
     \item \textbf{Implicit knowledge} inspired in YOLOR \cite{wang2021you}.
     \item \textbf{Exponential moving average} as the final inference model. 
 \end{itemize}

\subsection{Comparison with YOLOv4 and YOLOR}
In this section, we highlight the enhancements of YOLOv7 compared to previous YOLO models developed by the same authors.

Compared to YOLOv4, YOLOv7 achieved a 75\% reduction in parameters and a 36\% reduction in computation while simultaneously improving the average precision (AP) by 1.5\%.

In contrast to YOLOv4-tiny, YOLOv7-tiny managed to reduce parameters and computation by 39\% and 49\%, respectively, while maintaining the same AP.

Lastly, compared to YOLOR, YOLOv7 reduced the number of parameters and computation by 43\% and 15\%, respectively, along with a slight 0.4\% increase in AP.

Evaluated on MS COCO dataset test-dev 2017, YOLOv7-E6 achieved an AP of 55.9\% and AP\textsubscript{50} of 73.5\% with an input size of 1280 pixels with a speed of 50 FPS on an NVIDIA V100.

\section{DAMO-YOLO}
DAMO-YOLO \cite{xu2022damo} was published in ArXiv in November 2022 by Alibaba Group. Inspired by the current technologies, DAMO-YOLO included the following:

\begin{enumerate}
    \item \textbf{A Neural architecture search (NAS)}. They used a method called MAE-NAS \cite{tinynas} developed by Alibaba to find an efficient architecture automatically.  
    \item \textbf{A large neck}. Inspired by GiraffeDet \cite{tan2021giraffedet}, CSPNet \cite{wang2020cspnet}, and ELAN \cite{wang2022designing}, the authors designed a neck that can work in real-time called Efficient-RepGFPN.
    \item \textbf{A small head}. The authors found that a large neck and a small neck yield better performance, and they only left one linear layer for classification and one for regression. They called this approach ZeroHead.
    \item \textbf{AlignedOTA label assignment}. Dynamic label assignment methods, such as OTA\cite{ge2021ota} and TOOD\cite{feng2021tood}, have gained popularity due to their significant improvements over static methods. However, the misalignment between classification and regression remains a problem, partly because of the imbalance between classification and regression losses. To address this issue, their AlignOTA method introduces focal loss \cite{lin2017focal} into the classification cost and uses the IoU of prediction and ground truth box as the soft label, enabling the selection of aligned samples for each target and solving the problem from a global perspective.
    \item \textbf{Knowledge distillation}. Their proposed strategy consists of two stages: the teacher guiding the student in the first stage and the student fine-tuning independently in the second stage. Additionally, they incorporate two enhancements in the distillation approach: the Align Module, which adapts student features to the same resolution as the teacher's, and Channel-wise Dynamic Temperature, which normalizes teacher and student features to reduce the impact of real value differences.
\end{enumerate}

The authors generated scaled models named DAMO-YOLO-Tiny/Small/Medium, with the best model achieving an AP of 50.0 \% at 233 FPS on an NVIDIA V100.

\section{YOLOv8}
YOLOv8 \cite{Jocher_YOLO_by_Ultralytics_2023} was released in January 2023 by Ultralytics, the company that developed YOLOv5. YOLOv8 provided five scaled versions: YOLOv8n (nano), YOLOv8s (small), YOLOv8m (medium), YOLOv8l (large) and YOLOv8x (extra large). YOLOv8 supports multiple vision tasks such as object detection, segmentation, pose estimation, tracking, and classification.

\subsection{YOLOv8 Architecture}
\begin{figure}[ht]
  \centering
  \includegraphics[width=\linewidth]{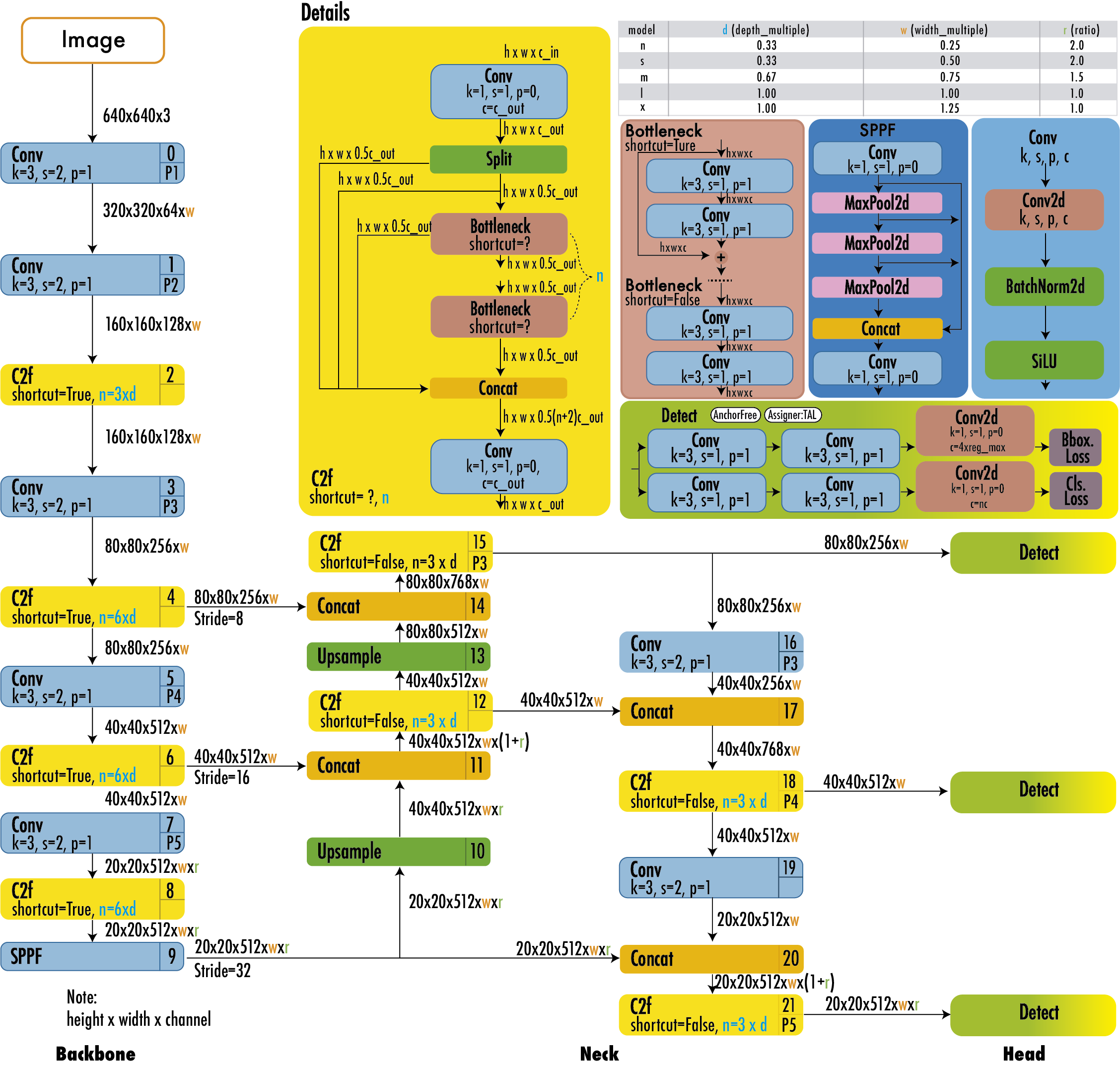}
  \caption{YOLOv8 Architecture. The architecture uses a modified CSPDarknet53 backbone. The C2f module replaces the CSPLayer used in YOLOv5. A spatial pyramid pooling fast (SPPF) layer accelerates computation by pooling features into a fixed-size map. Each convolution has batch normalization and SiLU activation. The head is decoupled to process objectness, classification, and regression tasks independently. Diagram based in \cite{mmyolo_vyolov8}.}
  \label{fig:yolov8_arch}
\end{figure}

Figure \ref{fig:yolov8_arch} shows the detailed architecture of YOLOv8. 
YOLOv8 uses a similar backbone as YOLOv5 with some changes on the CSPLayer, now called the C2f module. The C2f module (cross-stage partial bottleneck with two convolutions) combines high-level features with contextual information to improve detection accuracy. 

YOLOv8 uses an anchor-free model with a decoupled head to independently process objectness, classification, and regression tasks. This design allows each branch to focus on its task and improves the model's overall accuracy. In the output layer of YOLOv8, they used the sigmoid function as the activation function for the objectness score, representing the probability that the bounding box contains an object. It uses the softmax function for the class probabilities, representing the objects' probabilities belonging to each possible class. 

YOLOv8 uses CIoU \cite{zheng2020distance} and DFL \cite{li2020generalized} loss functions for bounding box loss and binary cross-entropy for classification loss. These losses have improved object detection performance, particularly when dealing with smaller objects.

YOLOv8 also provides a semantic segmentation model called YOLOv8-Seg model. The backbone is a CSPDarknet53 feature extractor, followed by a C2f module instead of the traditional YOLO neck architecture. The C2f module is followed by two segmentation heads, which learn to predict the semantic segmentation masks for the input image. The model has similar detection heads to YOLOv8, consisting of five detection modules and a prediction layer. The YOLOv8-Seg model has achieved state-of-the-art results on various object detection and semantic segmentation benchmarks while maintaining high speed and efficiency.


YOLOv8 can be run from the command line interface (CLI), or it can also be installed as a PIP package. In addition, it comes with multiple integrations for labeling, training, and deploying.

Evaluated on MS COCO dataset test-dev 2017, YOLOv8x achieved an AP of 53.9\% with an image size of 640 pixels (compared to 50.7\% of YOLOv5 on the same input size) with a speed of 280 FPS on an NVIDIA A100 and TensorRT.

\section{PP-YOLO, PP-YOLOv2, and PP-YOLOE}
PP-YOLO models have been growing parallel to the YOLO models we described. However, we decided to group them in a single section because they began with YOLOv3 and had been gradually improving upon the previous PP-YOLO version. Nevertheless, these models have been influential in the evolution of YOLO.  
PP-YOLO \cite{long2020pp} similar to YOLOv4 and YOLOv5 was based on YOLOv3. It was published in ArXiv in July 2020 by researchers from Baidu Inc. The authors used the PaddlePaddle \cite{ma2019paddlepaddle} deep learning platform, hence its \emph{PP} name. Following the trend we have seen starting with YOLOv4, PP-YOLO added ten existing tricks to improve the detector's accuracy, keeping the speed unchanged. According to the authors, this paper was not intended to introduce a novel object detector but to show how to build a better detector step by step.
Most of the tricks PP-YOLO uses are different from the ones used in YOLOv4, and the ones that overlap use a different implementation.
The changes of PP-YOLO concerning YOLOv3 are:

\begin{enumerate}
    \item \textbf{A ResNet50-vd backbone} replacing the DarkNet-53 backbone with an architecture augmented with deformable convolutions \cite{dai2017deformable} in the last stage and a distilled pre-trained model, which has a higher classification accuracy on ImageNet. This architecture was called ResNet5-vd-dcn.
    \item \textbf{A larger batch size} to improve training stability, they went from 64 to 192, along with an updated training schedule and learning rate.
    \item \textbf{Maintained moving averages} for the trained parameters and use them instead of the final trained values.
    \item \textbf{DropBlock} is applied only to the FPN.
    \item \textbf{An IoU loss} is added in another branch along with the L1-loss for bounding box regression.
    \item \textbf{An IoU prediction branch} is added to measure localization accuracy along with an IoU aware loss. During inference, YOLOv3 multiplies the classification probability and objectiveness score to compute the final detection, PP-YOLO also multiplies the predicted IoU to consider the localization accuracy.
    \item \textbf{Grid Sensitive approach} similar to YOLOv4 is used to improve the bounding box center prediction at the grid boundary.
    \item \textbf{Matrix NMS} \cite{xinlong2020solov2} is used, which can be run in parallel making it faster than traditional NMS. 
    \item \textbf{CoordConv} \cite{liu2018intriguing} is used for the $1\times1$ convolution of the FPN, and on the first convolution layer in the detection head. CoordConv allows the network to learn translational invariance improving the detection localization.
    \item \textbf{Spatial Pyramid Pooling} is used only on the top feature map to increase the receptive field of the backbone.
\end{enumerate}

\subsection{PP-YOLO augmentations and preprocessing}
PP-YOLO used the following augmentations and preprocessing:

\begin{enumerate}
    \item Mixup Training \cite{zhang2017mixup} with a weight sampled from $Beta(\alpha,\beta)$ distribution where $\alpha = 1.5$ and $\beta = 1.5$.
    \item Random Color Distortion.
    \item Random Expand.
    \item Random Crop and Random Flip with a probability of 0.5.
    \item RGB channel z-score normalization with a mean of $[0.485, 0.456, 0.406]$ and a standard deviation of $[0.229, 0.224, 0.225]$. 
    \item Multiple image sizes evenly drawn from [320, 352, 384, 416, 448, 480, 512, 544, 576, 608].
\end{enumerate}

Evaluated on MS COCO dataset test-dev 2017, PP-YOLO achieved an AP of 45.9\% and AP\textsubscript{50} of 65.2\% at 73 FPS on an NVIDIA V100.

\subsection{PP-YOLOv2}
PP-YOLOv2 \cite{huang2021pp} was published in ArXiv on April 2021 and added four refinements to PP-YOLO that increased performance from 45.9\% AP to 49.5\% AP at 69 FPS on NVIDIA V100. 
The changes of PP-YOLOv2 concerning PP-YOLO are the following:

\begin{enumerate}
    \item \textbf{Backbone changed from ResNet50 to ResNet101}. 
    \item \textbf{Path aggregation network (PAN)} instead of FPN similar to YOLOv4.
    \item \textbf{Mish Activation Function}. Unlike YOLOv4 and YOLOv5, they only applied the mish activation function in the detection neck to keep the backbone unchanged with ReLU.
    \item \textbf{Larger input sizes} help to increase performance on small objects. They expanded the largest input size from 608 to 768 and reduced the batch size from 24 to 12 images per GPU. The input sizes are evenly drawn from [320, 352, 384, 416, 448, 480, 512, 544, 576, 608, 640, 672, 704, 736, 768].
    \item \textbf{A modified IoU aware branch}. They modified the calculation of the IoU aware loss calculation using a soft label format instead of a soft weight format. 
\end{enumerate}

\subsection{PP-YOLOE}
\begin{figure}[ht]
  \centering
  \includegraphics[width=\linewidth]{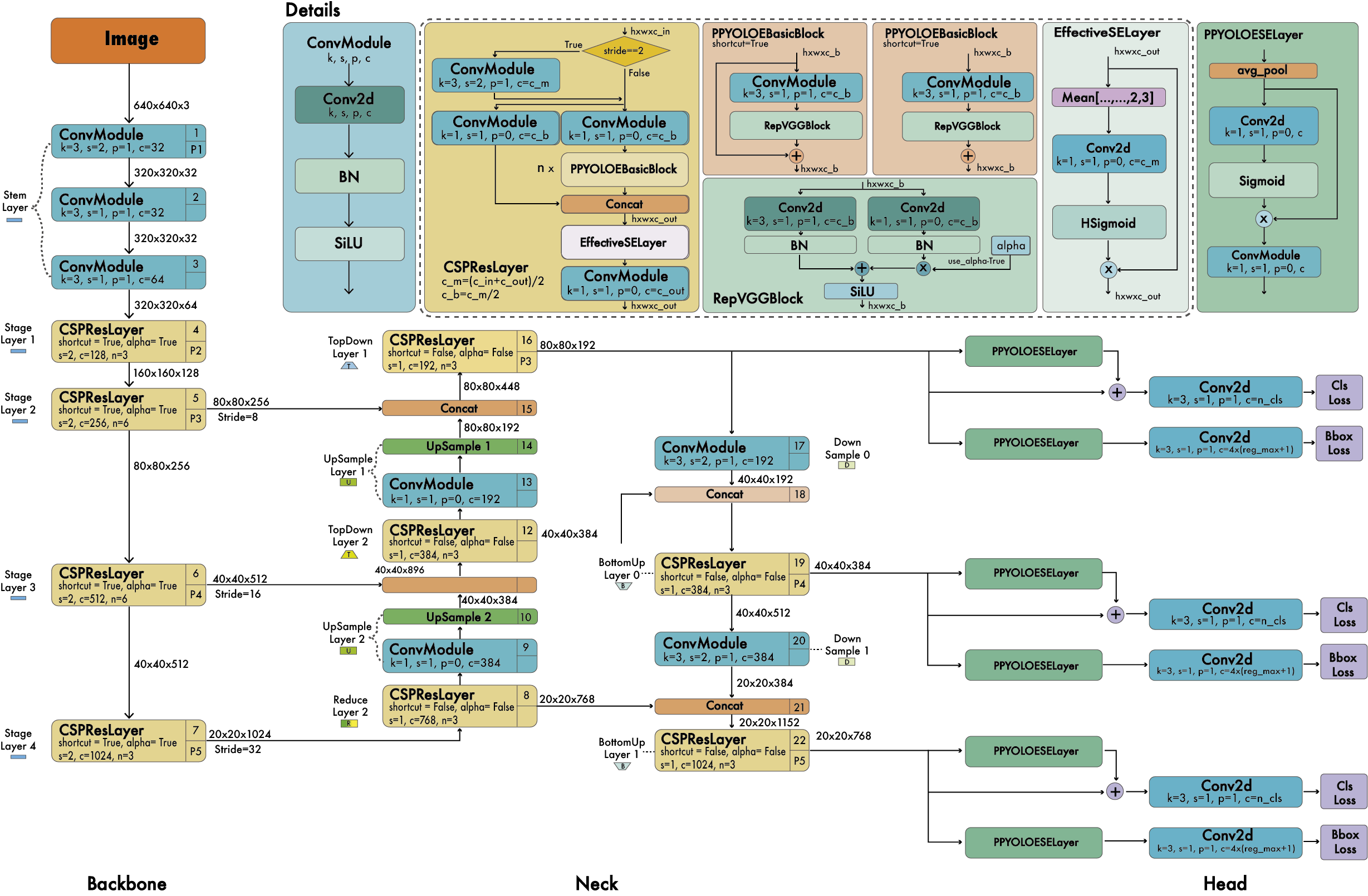}
  \caption{PP-YOLOE Architecture. The backbone is based on CSPRepResNet, the neck uses a path aggregation network, and the head uses ES layers to form an Efficient Task-aligned Head (ET-head). Diagram based in \cite{mmyolo_ppyoloe}.}
  \label{fig:ppyoloe_arch}
\end{figure}

PP-YOLOE \cite{xu2022pp} was published in ArXiv in March 2022. It added improvements upon PP-YOLOv2 achieving a performance of 51.4\% AP at 78.1 FPS on NVIDIA V100. Figure \ref{fig:ppyoloe_arch} shows a detailed architecture diagram.
The main changes of PP-YOLOE concerning PP-YOLOv2 are:
\begin{enumerate}
    \item \textbf{Anchor-free}. Following the time trends driven by the works of \cite{tian2019fcos,duan2019centernet,law2018cornernet,ge2021yolox}, PP-YOLOE uses an anchor-free architecture.
    \item \textbf{New backbone and neck}. Inspired by TreeNet \cite{rao2021treenet}, the authors modified the architecture of the backbone and neck with RepResBlocks combining residual and dense connections. 
    \item \textbf{Task Alignment Learning (TAL)}. YOLOX was the first to bring up the problem of task misalignment, where the classification confidence and the location accuracy do not agree in all cases. To reduce this problem, PP-YOLOE implemented TAL as proposed in TOOD \cite{feng2021tood}, which includes a dynamic label assignment combined with a task-alignment loss.
    \item \textbf{Efficient Task-aligned Head (ET-head)}. Different from YOLOX where the classification and locations heads were decoupled, PP-YOLOE instead used a single head based on TOOD to improve speed and accuracy.
    \item \textbf{Varifocal (VFL) and Distribution focal loss (DFL)}. VFL \cite{zhang2021varifocalnet} weights loss of positive samples using target score, giving higher weight to those with high IoU. This prioritizes high-quality samples during training. Similarly, both use IoU-aware classification score (IACS) as the target, allowing for joint learning of classification and localization quality, leading to consistency between training and inference. On the other hand, DFL \cite{li2020generalized} extends Focal Loss from discrete to continuous labels, enabling successful optimization of improved representations that combine quality estimation and class prediction. This allows for an accurate depiction of flexible distribution in real data, eliminating the risk of inconsistency.
\end{enumerate}

Like previous YOLO versions, the authors generated multiple scaled models by varying the width and depth of the backbone and neck. The models are called PP-YOLOE-s (small), PP-YOLOE-m (medium), PP-YOLOE-l (large), and PP-YOLOE-x (extra large).




\section{YOLO-NAS}
YOLO-NAS \cite{YOLO_NAS_by_Deci_2023} was released in May 2023 by Deci, a company that develops production-grade models and tools to build, optimize, and deploy deep learning models.
YOLO-NAS is designed to detect small objects, improve localization accuracy, and enhance the performance-per-compute ratio, making it suitable for real-time edge-device applications. In addition, its open-source architecture is available for research use.

The novelty of YOLO-NAS includes the following:
\begin{itemize}
    \item Quantization aware modules \cite{chu2022make} called QSP and QCI that combine re-parameterization for 8-bit quantization to minimize the accuracy loss during post-training quantization. 
    \item Automatic architecture design using AutoNAC, Deci's proprietary NAS technology. 
    \item Hybrid quantization method to selectively quantize certain parts of a model to balance latency and accuracy instead of standard quantization, where all the layers are affected.
    \item A pre-training regimen with automatically labeled data, self-distillation, and large datasets.
\end{itemize}

\begin{figure}[ht]
  \centering
  \includegraphics[width=\linewidth]{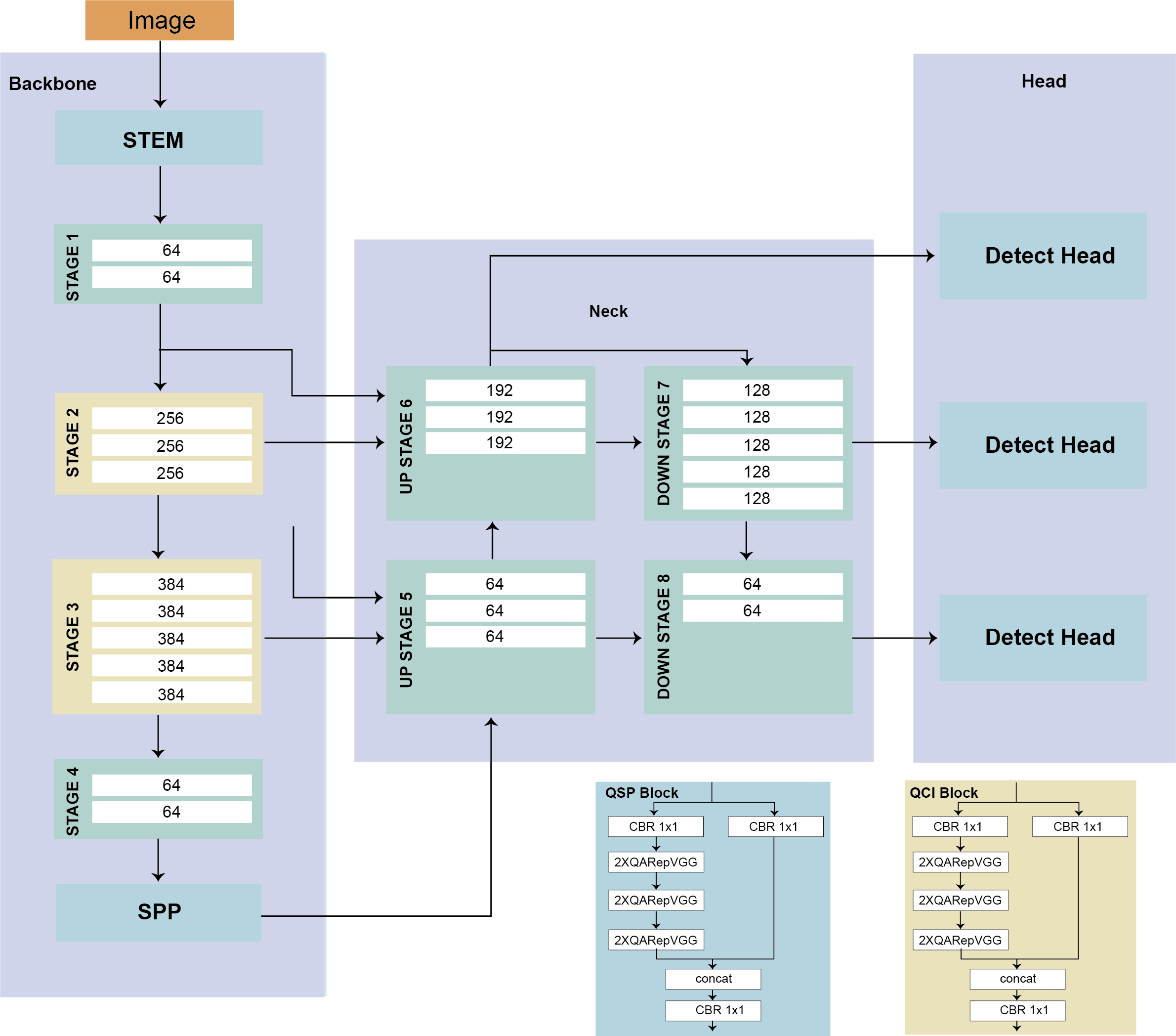}
  \caption{YOLO-NAS Architecture. The architecture is found automatically via a Neural Architecture Search (NAS) system called AutoNAC to balance latency vs. throughput. They generated three architectures called YOLO-NASS (small), YOLO-NASM (medium), and YOLO-NASL (large), varying the depth and positions of the QSP and QCI blocks. The figure shows the YOLO-NASL architecture.}
  \label{fig:yolo_nas_arch}
\end{figure}

The AutoNAC system, which was instrumental in creating YOLO-NAS, is versatile and can accommodate any task, the specifics of the data, the environment for making inferences, and the setting of performance goals. It assists users in identifying the most suitable structure that offers the perfect blend of precision and inference speed for their particular use. This technology considers the data and hardware and other elements involved in the inference process, such as compilers and quantization. In addition, RepVGG blocks were incorporated into the model architecture during the NAS process for compatibility with Post-Training Quantization (PTQ). They generated three architectures by varying the depth and positions of the QSP and QCI blocks: YOLO-NASS, YOLO-NASM, and YOLO-NASL (S,M,L for small, medium, and large, respectively). Figure \ref{fig:yolo_nas_arch} shows the model architecture for YOLO-NASL.

The model is pre-trained on Objects365 \cite{shao2019objects365}, which contains two million images and 365 categories, then the COCO dataset was used to generate pseudo-labels. Finally, the models are trained with the original 118k train images of the COCO dataset.

At this writing, three YOLO-NAS models have been released in FP32, FP16, and INT8 precisions, achieving an AP of 52.2\% on MS COCO with 16-bit precision.

\section{YOLO with Transformers}
With the rise of the Transformer~\cite{vaswani2017attention} taking over most Deep Learning tasks from Language and Audio Processing to Vision, it was natural for Transformers and YOLO to be combined. One of the first attempts at using transformers for object detection was You Only Look at One Sequence or YOLOS~\cite{fang2021you}, turned a pre-trained Vision Transfomer (ViT)~\cite{dosovitskiy2020image} from image classification to object detection, achieving 42.0 \% AP on MS COCO dataset. The changes made to ViT were two: 1) replace one [CLS] token used in classification with one hundred [DET] tokens for detection, and 2) replace the image classification loss in ViT with a bipartite matching loss similar to the End-to-end object detection with transformers~\cite{carion2020end}.

\begin{figure}[ht]
  \centering
  \includegraphics[width=13cm]{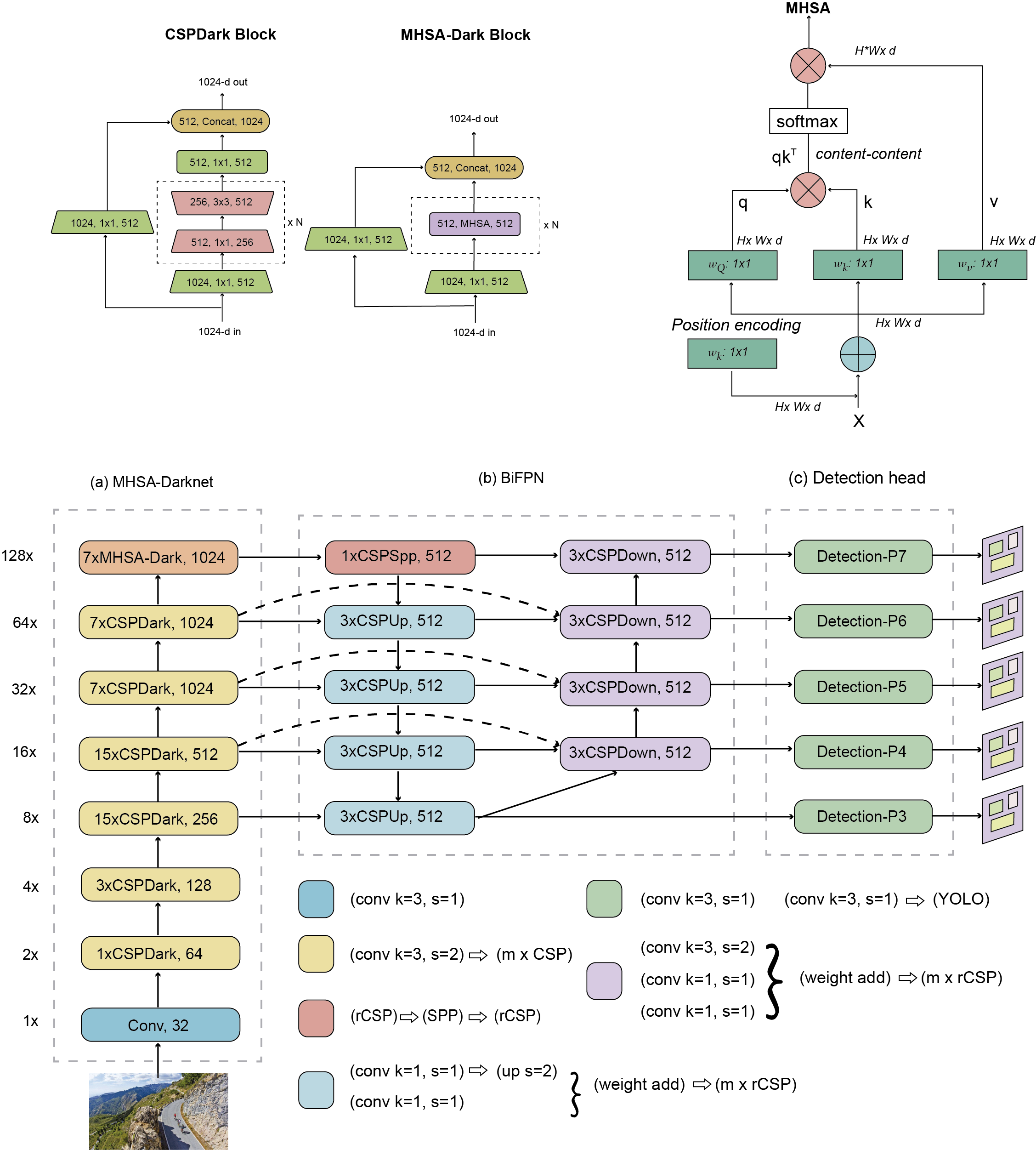}
  \caption{ViT-YOLO Architecture. The backbone MHSA-Darknet combines multi-head self-attention blocks (MHSA-Dark Block) with Cross-Stage Partial connection blocks (CSPDark block). The neck uses BiFPN to aggregate features from different backbone levels, and the head comprises five multi-scale detection heads. }
  \label{fig:vit_yolo_arch}
\end{figure}

Many works have combined transformers with YOLO-related architectures tailored to specific applications. For example, Zhang et al.~\cite{zhang2021vit}, motivated by the robustness of Vision Transformers to occlusions, perturbations, and domain shifts, proposed ViT-YOLO, a hybrid architecture that combines CSP-Darknet~\cite{bochkovskiy2020yolov4} and multi-head self-attention (MHSA-Darknet) in the backbone along with bidirectional feature pyramid networks (BiFPN)~\cite{tan2020efficientdet} for the neck and multi-scale detection heads like YOLOv3. Their specific use case was for object detection in drone images. Figure~\ref{fig:vit_yolo_arch} shows the detailed architecture of ViT-YOLO.  

MSFT-YOLO~\cite{guo2022msft} adds transformer-based modules to the backbone and detection heads intending to detect defects on the steel surface. NRT-YOLO~\cite{liu2022nrt} (Nested Residual Transformer) tries to address the problem of tiny objects in remote sensing images. Adding an extra prediction head, feature fusion layers, and a residual transformer module, NRT-YOLO improved YOLOv5l by 5.4\% in the DOTA data set~\cite{xia2018dota}.



In remote sensing applications, YOLO-SD~\cite{wang2022yolosd} tried to improve the detection accuracy for small ships in synthetic aperture radar (SAR) images. They started with YOLOX~\cite{ge2021yolox} coupled with multi-scale convolution (MSC) to improve the detection at different scales and feature transformer modules to capture global features. The authors showed that these changes improved the accuracy of YOLO-SD compared with YOLOX in the HRSID dataset~\cite{wei2020hrsid}.



Another interesting attempt to combine YOLO with detection transformer (DETR)~\cite{carion2020end} is the case of DEYO~\cite{ouyang2022deyo} comprising two stages: a YOLOv5-based model followed by a DETR-like model. The first stage generates high-quality query and anchors that input to the second stage. The results show a faster convergence time and better performance than DETR, achieving 52.1\% AP in the COCO detection benchmark.


\section{Discussion}
This paper examined 16 YOLO versions, ranging from the original YOLO model to the most recent YOLO-NAS. Table \ref{tab:summary} provides an overview of the YOLO versions discussed. From this table, we can identify several key patterns:

\begin{itemize}
    \item  \textbf{Anchors}: The original YOLO model was relatively simple and did not employ anchors, while the state-of-the-art relied on two-stage detectors with anchors. YOLOv2 incorporated anchors, leading to improvements in bounding box prediction accuracy. This trend persisted for five years until YOLOX introduced an anchor-less approach that achieved state-of-the-art results. Since then, subsequent YOLO versions have abandoned the use of anchors.
    \item \textbf{Framework}: Initially, YOLO was developed using the Darknet framework, with subsequent versions following suit. However, when Ultralytics ported YOLOv3 to PyTorch, the remaining YOLO versions were developed using PyTorch, leading to a surge in enhancements. Another deep learning language utilized is PaddlePaddle, an open-source framework initially developed by Baidu.
    \item \textbf{Backbone}: The backbone architectures of YOLO models have undergone significant changes over time. Starting with the Darknet architecture, which comprised simple convolutional and max pooling layers, later models incorporated cross-stage partial connections (CSP) in YOLOv4, reparameterization in YOLOv6 and YOLOv7, and neural architecture search in DAMO-YOLO and YOLO-NAS.
    \item \textbf{Performance}: While the performance of YOLO models has improved over time, it is worth noting that they often prioritize balancing speed and accuracy rather than solely focusing on accuracy. This tradeoff is essential to the YOLO framework, allowing for real-time object detection across various applications.
\end{itemize}

\begin{table}
  \centering
  \caption{Summary of YOLO architectures. The metric reported for YOLO and YOLOv2 were on VOC2007, while the rest are reported on COCO2017. The NAS-YOLO model reported has 16-bit precision. }
  \label{tab:summary}
  \begin{tabular}{lllllll}
   \hline
    Version & Date  & Anchor & Framework & Backbone  & AP (\%) \\ 
    \hline
    YOLO        & 2015  & No    & Darknet  & Darknet24 & \textit{63.4} \\
    YOLOv2      & 2016  & Yes   & Darknet  & Darknet24 & \textit{78.6} \\
    YOLOv3      & 2018  & Yes   & Darknet  & Darknet53 & $33.0$ \\
    YOLOv4      & 2020  & Yes   & Darknet  & CSPDarknet53 & $43.5$ \\
    YOLOv5      & 2020  & Yes   & Pytorch  & YOLOv5CSPDarknet & $55.8$ \\
    PP-YOLO     & 2020  & Yes   & PaddlePaddle & ResNet50-vd & $45.9$ \\
    Scaled-YOLOv4 & 2021 & Yes  & Pytorch  & CSPDarknet & $56.0$ \\
    PP-YOLOv2   & 2021  & Yes   & PaddlePaddle & ResNet101-vd & $50.3$ \\
    YOLOR       & 2021  & Yes   & Pytorch  & CSPDarknet & $55.4$ \\
    YOLOX       & 2021  & No    & Pytorch  & YOLOXCSPDarknet & $51.2$ \\
    PP-YOLOE    & 2022  & No    & PaddlePaddle & CSPRepResNet & $54.7$ \\
    YOLOv6      & 2022  & No    & Pytorch  & EfficientRep & $52.5$ \\
    YOLOv7      & 2022  & No    & Pytorch  & YOLOv7Backbone & $56.8$ \\ 
    DAMO-YOLO  & 2022  & No   & Pytorch  & MAE-NAS & $50.0$ \\
    YOLOv8      & 2023  & No    & Pytorch  & YOLOv8CSPDarknet  & $53.9$ \\
    YOLO-NAS   & 2023  & No    & Pytorch  & NAS & $52.2$ \\
\hline
\end{tabular}
\end{table}

\subsection{Tradeoff between speed and accuracy}
The YOLO family of object detection models has consistently focused on balancing speed and accuracy, aiming to deliver real-time performance without sacrificing the quality of detection results. As the YOLO framework has evolved through its various iterations, this tradeoff has been a recurring theme, with each version seeking to optimize these competing objectives differently.
In the original YOLO model, the primary focus was on achieving high-speed object detection. The model utilized a single convolutional neural network (CNN) to directly predict object locations and classes from the input image, enabling real-time processing. However, this emphasis on speed led to a compromise in accuracy, mainly when dealing with small objects or objects with overlapping bounding boxes.

Subsequent YOLO versions introduced refinements and enhancements to address these limitations while maintaining the framework's real-time capabilities. For instance, YOLOv2 (YOLO9000) introduced anchor boxes and passthrough layers to improve the localization of objects, resulting in higher accuracy. In addition, YOLOv3 enhanced the model's performance by employing a multi-scale feature extraction architecture, allowing for better object detection across various scales.

The tradeoff between speed and accuracy became more nuanced as the YOLO framework evolved. Models like YOLOv4 and YOLOv5 introduced innovations, such as new network backbones, improved data augmentation techniques, and optimized training strategies. These developments led to significant gains in accuracy without drastically affecting the models' real-time performance. 

\begin{figure}[H]
  \includegraphics[width=\textwidth]{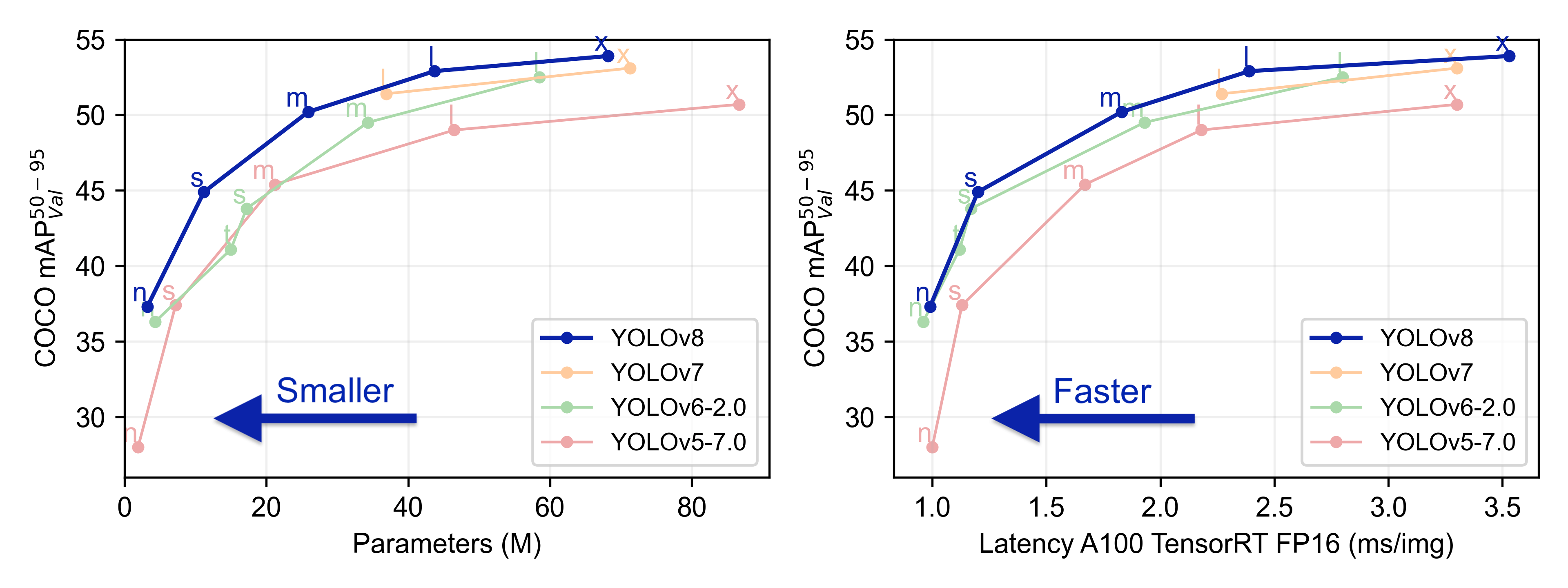}
  \caption{Performance comparison of YOLO object detection models. The left plot illustrates the relationship between model complexity (measured by the number of parameters) and detection accuracy (COCO mAP50-95). The right plot shows the tradeoff between inference speed (latency on A100 TensorRT FP16) and accuracy for the same models. Each model version is represented by a distinct color, with markers indicating size variants from \emph{nano} to \emph{extra}. Plots taken from~\cite{Jocher_YOLO_v8}.}
  \label{fig:yolo-comparison-plots}
\end{figure}

From YOLOv5, all official YOLO models have fine-tuned the tradeoff between speed and accuracy,  offering different model scales to suit specific applications and hardware requirements. For instance, these versions often provide lightweight models optimized for edge devices, trading accuracy for reduced computational complexity and faster processing times. Figure~\ref{fig:yolo-comparison-plots}~\cite{Jocher_YOLO_v8} shows the comparison of the different model scales from YOLOv5 to YOLOv8. The figure presents a comparative analysis of different versions of YOLO models in terms of their complexity and performance. The left graph plots the number of parameters (in millions) against the mean average precision (mAP) on the COCO validation set, ranging from IOU thresholds of 50 to 95. It illustrates a clear trend where an increase in the number of parameters enhances the model's accuracy. Each model includes various scales indicated by \emph{n} (nano),  \emph{s} (small), \emph{m} (medium), \emph{l} (large), and \emph{x} (extra-large).

The right graph contrasts the inference latency on an NVIDIA A100 GPU, utilizing TensorRT FP16, with the same mAP performance metric. Here, the tradeoff between the inference speed and the detection accuracy is evident. Lower latency values, indicating faster model inference, typically result in reduced accuracy. Conversely, models with higher latency tend to achieve better performance on the COCO mAP metric. This relationship is pivotal for applications where real-time processing is crucial, and the choice of model is influenced by the requirement to balance speed and accuracy.

\section{The future of YOLO}






As the YOLO framework continues to evolve, we anticipate that the following trends and possibilities will shape future developments:

\textbf{Incorporation of Latest Techniques}. Researchers and developers will continue to refine the YOLO architecture by leveraging state-of-the-art methods in deep learning, data augmentation, and training techniques. This ongoing innovation will likely improve the model's performance, robustness, and efficiency.

\textbf{Benchmark Evolution}. The current benchmark for evaluating object detection models, COCO 2017, may eventually be replaced by a more advanced and challenging benchmark. This mirrors the transition from the VOC 2007 benchmark used in the first two YOLO versions, reflecting the need for more demanding benchmarks as models grow more sophisticated and accurate.

\textbf{Proliferation of YOLO Models and Applications}. As the YOLO framework progresses, we expect to witness an increase in the number of YOLO models released each year, along with a corresponding expansion of applications. As the framework becomes more versatile and powerful, it will likely be employed in more varied domains, from home appliances devices to autonomous cars.

\textbf{Expansion into New Domains}. YOLO models have the potential to expand beyond object detection and segmentation, exploring domains such as object tracking in videos and 3D keypoint estimation. We anticipate YOLO models to transition into multi-modal frameworks, incorporating both vision and language, video, and sound processing. As these models evolve, they may serve as the foundation for innovative solutions catering to a broader spectrum of computer vision and multimedia tasks.

\textbf{Adaptability to Diverse Hardware}. YOLO models will further span hardware platforms, from IoT devices to high-performance computing clusters. This adaptability will enable deploying YOLO models in various contexts, depending on the application's requirements and constraints. In addition, by tailoring the models to suit different hardware specifications, YOLO can be made accessible and effective for more users and industries.



\section{Acknowledgments}
We thank the National Council for Science and Technology (CONACYT) for its support through the National Research System (SNI).

\bibliographystyle{ieeetr}
\bibliography{references}  

\end{document}